\documentclass[10pt,twocolumn,letterpaper]{article}
\usepackage{bm}
\usepackage{booktabs}
\usepackage{makecell} 
\usepackage{multirow}
\usepackage{adjustbox}
\usepackage{pifont}   
\usepackage{arydshln}
\newcommand{\cmark}{\ding{51}} %
\newcommand{\xmark}{\ding{55}} %
\usepackage{float}
\usepackage[table]{xcolor}

\usepackage{hhline}

\usepackage{cvpr}              
\definecolor{cvprblue}{rgb}{0.21,0.49,0.74}
\usepackage[pagebackref,breaklinks,colorlinks,allcolors=cvprblue]{hyperref}

\def\paperID{*****} 
\def\confName{CVPR}
\def\confYear{2026}

\title{Leave No Stone Unturned: Uncovering Holistic Audio-Visual Intrinsic Coherence for Deepfake Detection}

\author{Jielun Peng \quad 
Yabin Wang \quad
Yaqi Li \quad
Long Kong \quad
Xiaopeng Hong\thanks{Corresponding author} \\
Harbin Institute of Technology \\
{\tt\small 25s003052@stu.hit.edu.cn \quad wang-yabin@outlook.com \quad hongxiaopeng@ieee.org}
}

\begin{document}
\maketitle
\begin{abstract}
The rapid progress of generative AI has enabled hyper-realistic audio-visual deepfakes, intensifying threats to personal security and social trust. 
Most existing deepfake detectors rely either on uni-modal artifacts or audio-visual discrepancies, failing to jointly leverage both sources of information.
Moreover, detectors that rely on generator-specific artifacts tend to exhibit degraded generalization when confronted with unseen forgeries. 
We argue that robust and generalizable detection should be grounded in intrinsic audio-visual coherence within and across modalities. Accordingly, we propose \textit{HAVIC}, a Holistic Audio-Visual Intrinsic Coherence-based deepfake detector. 
HAVIC first learns priors of modality-specific structural coherence, inter-modal micro- and macro-coherence by pre-training on authentic videos. 
Based on the learned priors, HAVIC further performs holistic adaptive aggregation to dynamically fuse audio-visual features for deepfake detection. 
Additionally, we introduce \textit{HiFi-AVDF}, a high-fidelity audio-visual deepfake dataset featuring both text-to-video and image-to-video forgeries from state-of-the-art commercial generators. 
Extensive experiments across several benchmarks demonstrate that HAVIC significantly outperforms existing state-of-the-art methods, achieving improvements of 9.39\% AP and 9.37\% AUC on the most challenging cross-dataset scenario.  
Our code and dataset are available at \url{https://github.com/tuffy-studio/HAVIC}.
\end{abstract}    
\vspace{-1em}
\section{Introduction}
\label{sec:intro}

The rapid advancement of generative AI has spurred the creation of hyper-realistic deepfakes across both visual and audio modalities. 
While these technologies hold creative potential, their malicious use for disinformation, fraud, and harassment poses a significant threat to personal security and social trust. 
Consequently, developing reliable deepfake detection methods has become increasingly important.

Early deepfake detectors~\cite{cViTforDFD,Multi_attentional_deepfake_detection,high-frequency,lip-forensics} typically focus on intra-modal artifacts and are trained to identify low-level fingerprints or distortions left by specific generators~\cite{dflip}, such as periodic frequency signals~\cite{frequency-aware,qian2020thinking}, visual texture patterns~\cite{Xception,dang2020detection}, and spatial-temporal inconsistency~\cite{ftcn,altfreezing,peng2024deepfakes}. 
While competitive on early benchmarks, artifact-centric detectors are brittle under cross-generator shift~\cite{liu2024turns,li2025survey}, transferring poorly to state-of-the-art generators such as DALL·E~\cite{DALL-E} and Sora~\cite{sora2}. 
Some methods leverage pre-trained models~\cite{opensdi,effort,poundnet} or perform pre-training on real data~\cite{fsfm,marlin} to incorporate prior knowledge, thereby mitigating overfitting to specific fake patterns and improving generalization. 
However, relying only on intra-modal cues limits their effectiveness in multi-modal manipulation scenarios. 
Stepping up from single-modality flaws, some works scrutinize inter-modal inconsistencies~\cite{AVgraph,avoid-df,avprompt,Emotionsdon'tlie, oorloff2024avff}, such as the mismatch between lip movements and speech~\cite{speechforensics,phoneme_viseme_mismatch,avh-align}. 
Nevertheless, as lip-sync generation technologies become increasingly mature, this line of defense is also being systematically dismantled~\cite{lipfd,pia}.


We argue that detectors with strong robustness and cross-generator generalization should, leaving no stone unturned, ground their decisions in real-world structural, temporal, and semantic coherence rather than relying on any single modality or specific artifact.
We summarize coherence into three levels.
The first level is \textit{modality-specific structural coherence}, which examines the structural plausibility within a single modality. It targets flaws where objects or sounds defy common-sense physics, such as unnatural deformities and acoustic distortions.
The second level is \textit{inter-modal micro-coherence}, which requires precise segment-level alignment between audio and video streams.
Its violation is evident in the desynchronization between phoneme articulation and its corresponding lip shape. 
Finally, the highest level is \textit{inter-modal macro-coherence}, which captures instance-level consistency between audio and video modalities.  
A common flaw occurs when the overall scene conveyed by the audio does not match the visual content, such as hearing a person speaking about one action while the video shows a completely unrelated activity.
Therefore, a truly generalizable detector must adopt a holistic approach, evaluating authenticity by integrating the structural, temporal, and semantic coherence.\looseness=-1

\begin{table}[t]
\centering
\resizebox{0.95\linewidth}{!}{
\begin{tabular}{l|c|c|c|c|c}
\hline
Dataset & Modality & T2V & I2V & Person \# & Year \\ \hline
FaceForensics++~\cite{Xception} & V  & \textcolor{red}{\ding{55}} & \textcolor{red}{\ding{55}} & -- & 2019 \\ 
WildDeepfake~\cite{wilddeepfake}    & V  & \textcolor{red}{\ding{55}} & \textcolor{red}{\ding{55}} & -- & 2020 \\ 
KoDF~\cite{kodf}                     & V  & \textcolor{red}{\ding{55}} & \textcolor{red}{\ding{55}} & 403 & 2021 \\ 
DF-Platter~\cite{df-platter}         & V  & \textcolor{red}{\ding{55}} & \textcolor{red}{\ding{55}} & 454 & 2023 \\ 
FakeAVCeleb~\cite{fakeavceleb}       & AV & \textcolor{red}{\ding{55}} & \textcolor{red}{\ding{55}} & 500 & 2021 \\ 
DefakeAVMiT~\cite{avoid-df}          & AV & \textcolor{red}{\ding{55}} & \textcolor{red}{\ding{55}} & 86  & 2023 \\ 
AV-Deepfake1M~\cite{avdeepfake1m}    & AV & \textcolor{red}{\ding{55}} & \textcolor{red}{\ding{55}} & 2,068 & 2024 \\
\hline
\textbf{HiFi-AVDF (Ours)}            & AV & \textcolor{green}{\ding{51}} & \textcolor{green}{\ding{51}} & \textbf{1,905} & \textbf{2025} \\ \hline
\end{tabular}
}
\vspace{-0.5em}
\caption{Comparison of representative deepfake datasets. 
Columns indicate manipulated modality (V: visual-only, AV: audio-visual), support for Text-to-Video (T2V) and Image-to-Video (I2V) generation, number of person, and publication year. 
Our HiFi-AVDF dataset contains forged content in both audio and visual modalities, and supports modern T2V and I2V generation.}
\label{tab:dataset_summary}
\vspace{-1.5em}
\end{table}

Moreover, existing benchmarks~\cite{fakeavceleb,kodf,Xception,df-platter,avoid-df,df-platter} are almost exclusively centered on legacy forgeries from early GANs~\cite{GAN,proGAN,styleGAN}, face-swapping~\cite{faceswap,faceshifter}, and audio-driven~\cite{talklip,wav2lip} methods, making them unreliable for assessing generalization against the newest wave of generative AI.
To rectify this, we present the \textit{High-Fidelity Audio-Visual DeepFake} (HiFi-AVDF) dataset, a high-quality benchmark tailored to assess generalization against recent generative models. Its pioneering contribution is the inclusion of synthesized videos from cutting-edge commercial Text-to-Video (T2V) and Image-to-Video (I2V) generators, 
including Veo 3.1~\cite{veo3.1}, Kling 2.5~\cite{kling2.5}, Seedance 1.0~\cite{seedance}, Pixverse V5~\cite{pixverse}, WAN2.5~\cite{wan2.5}, and Sora 2~\cite{sora2}. 
Crucially, in constructing the dataset, human experts actively prompted and screened for high-fidelity videos across diverse identities and scenarios.
As Tab.~\ref{tab:dataset_summary} shows, HiFi-AVDF includes both T2V and I2V from modern generators across 1,905 person, faithfully reflecting in-the-wild conditions.

To overcome the above limitations, we propose a noval \textbf{H}olistic \textbf{A}udio-\textbf{V}isual \textbf{I}ntrinsic \textbf{C}oherence-based deepfake Detector (\textbf{HAVIC}), which models holistic coherence within and across modalities.
HAVIC first undergoes a \textit{Holistic Coherence Priors} pre-training phase on large-scale authentic videos, where we adapt masked autoencoding to both audio and visual inputs and jointly optimize three self-supervised objectives: a \textit{Modality-Specific Hierarchical Reconstruction Loss} that learns comprehensive modality-specific structural coherence by reconstructing masked tokens from hierarchical encoder features, a \textit{Fine-grained Audio-Visual Contrastive Loss} that partitions high-level audio and visual features into temporal segments and performs contrastive learning with a soft negative pairs strategy to capture temporal alignment and inter-modal micro-coherence, and a \textit{Cross-modal Semantic Reconstruction Loss} that decodes the global semantic representations of one modality from the other to enforce inter-modal macro-coherence. After pre-training, the model with learned holistic coherence priors is used in \textit{Holistic Adaptive Aggregation} Classification phase, where an \textit{Adaptive Feature Aggregation module} learns to weigh hierarchical uni-modal features together with interaction-aware features, enabling the classifier to adaptively exploit low-level artifacts, high-level semantics, and cross-modal discrepancies for robust deepfake detection.
In summary, our main contributions are:

\begin{enumerate}
    \item We propose HAVIC, a novel two-stage framework for robust deepfake detection that first learns holistic audio-visual intrinsic coherence priors, and then adaptively aggregation audio-visual features for classification.
    \item We introduce the HiFi-AVDF dataset, a new and challenging benchmark featuring high-fidelity forgeries from state-of-the-art commercial generation models.
    \item Our framework achieves superior performance across challenging benchmarks, consistently outperforming existing state-of-the-art methods, demonstrating its effectiveness and generalizability.
\end{enumerate}


\section{Related Works}
\label{sec:rw}
\begin{figure*}[t]
    \centering
    \includegraphics[width=1.\linewidth]{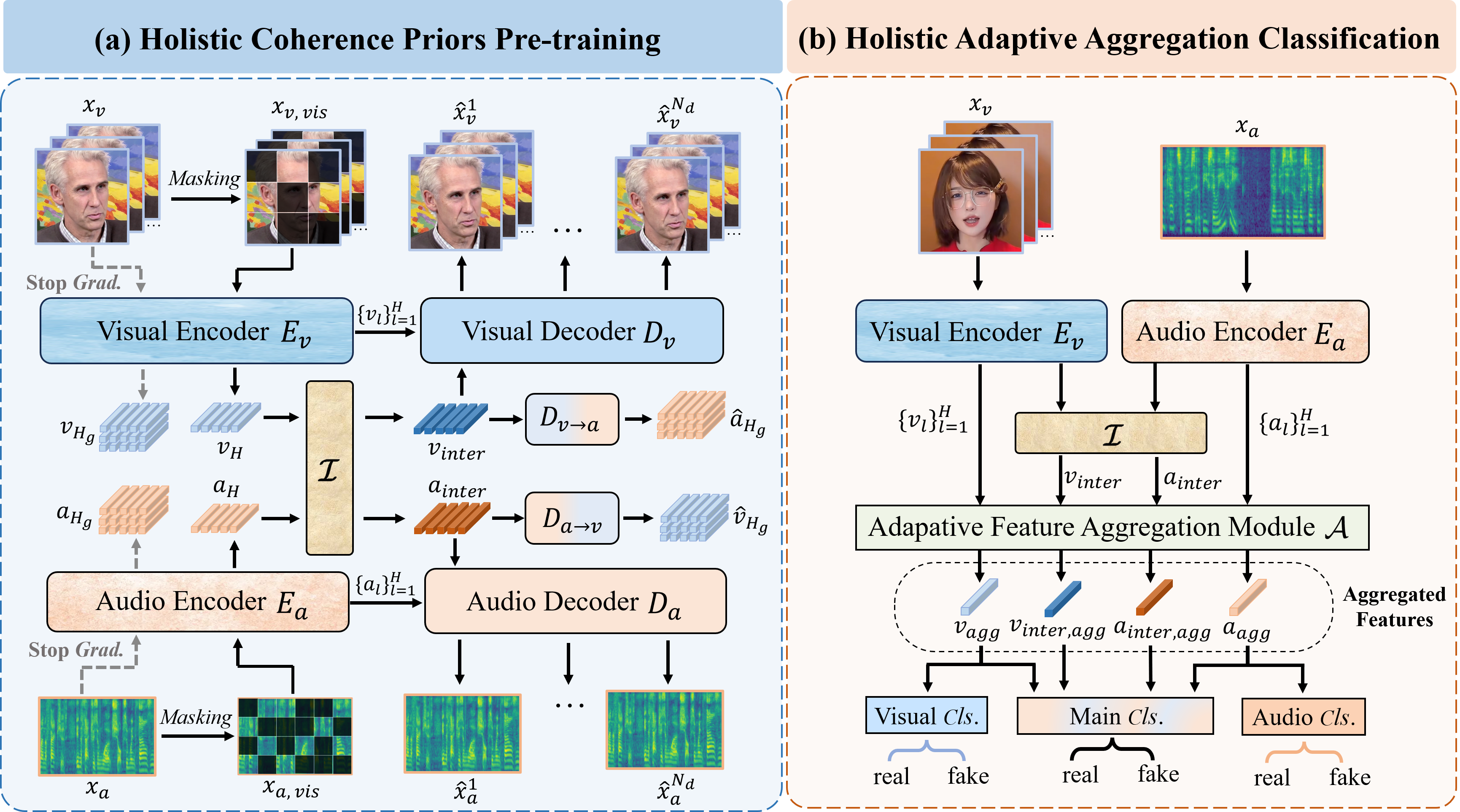}
    \vspace{-1.5em}
    \caption{\textbf{Overview of our proposed HAVIC.}
    \textbf{(a) Holistic Coherence Priors Pre-training phase}. The visible tokens $x_{v,vis}$ and $x_{a,vis}$ are encoded by $E_v$ and $E_a$ to produce hierarchical features $\{\bm{v}_l\}_{l=1}^H$ and $\{\bm{a}_l\}_{l=1}^H$. 
    The high-level representations $\bm{v}_H$ and $\bm{a}_H$ are further processed by the Audio-Visual Interaction Module $\mathcal{I}$ to yield interaction-aware features 
    $\bm{v}_{inter}$ and $\bm{a}_{inter}$. 
    These features are decoded by modality-specific decoders $D_v$ and $D_a$, where each layer integrates hierarchical features for input reconstruction. 
    In parallel, $\bm{v}_{inter}$ and $\bm{a}_{inter}$ are fed into cross-modal decoders $D_{v\to a}$ and $D_{a\to v}$ to reconstruct the counterpart semantics $\hat{\bm{a}}_{H_g}$ and $\hat{\bm{v}}_{H_g}$, supervised by gradient-stopped targets 
    $\bm{a}_{H_g}$ and $\bm{v}_{H_g}$.
    \textbf{(b) Holistic Adaptive Aggregation Classification phase}. The pre-trained $E_v$, $E_a$, and $\mathcal{I}$ with learned holistic coherence priors are used to extract features from the input sample. These features are aggregated by the Adaptive Feature Aggregation module $\mathcal{A}$ and then fed into the classifiers to predict both modality-specific and overall authenticity.}
    \label{fig:pipeline_rl}
    \vspace{-1em}
\end{figure*}

\subsection{Audio-Visual Self-Supervised Learning}
Self-supervised learning (SSL) aims to learn meaningful representations from unlabeled data by solving pretext tasks such as predicting masked parts~\cite{bert,avhubert,MAE,v-jepa} or contrasting positive and negative pairs~\cite{infonce,supervised-contrastive-learning,simCLR,clip}.
Recently, contrastive learning~\cite{infonce} and Masked Autoencoders (MAE)~\cite{MAE} have become the two primary paradigms in SSL.
Representative works include CLIP~\cite{clip}, which aligns visual and textual modalities through large-scale contrastive learning. AudioMAE~\cite{audiomae} and VideoMAE~\cite{tong2022videomae} extend the MAE framework to audio and video domains, respectively.
FSFM~\cite{fsfm} further integrates MAE and instance discrimination to learn universal facial representations. 
To exploit cross-modal synergy, audio-visual SSL frameworks have emerged~\cite{avf-mae++,crossmae,mavil}. CAV-MAE~\cite{cav-mae} unifies contrastive learning and MAE, revealing their complementarity in joint audio-visual representation learning. HiCMAE~\cite{hicmae} further employs hierarchical contrastive and reconstruction objectives for audio-visual emotion recognition tasks. CAV-MAE Sync~\cite{cav-mae-sync} improves temporal granularity between visual and audio frames, supporting downstream tasks such as audio-visual classification and localization.

\subsection{Deepfake Detection}
\noindent\textbf{Visual Deepfake Detection.}
Methods leveraging visual artifacts span both image-~\cite{Xception,gu2022exploiting} and video-based~\cite{ftcn,yan2025generalizing} settings, where image-based approaches focus on spatial inconsistencies (e.g., abnormal facial regions~\cite{dang2020detection,Multi_attentional_deepfake_detection}), while video-based methods extend the analysis to capture both spatial and temporal inconsistencies, such as irregular mouth movements~\cite{lip-forensics}.
Recent studies have explored various strategies to improve generalization to unseen forgeries. One common approach is to augment training data with synthetic samples, promoting the learning of generic forgery representations. Representative methods include artifact simulation, e.g., blending boundaries~\cite{face_x_ray,SBI,sladd}, and latent-space augmentation to enhance diversity~\cite{LSDA}.
Another effective direction is leveraging the pre-trained knowledge to alleviate overfitting. For example, MARLIN \cite{marlin} adopts an enhanced VideoMAE~\cite{tong2022videomae} strategy to learn general representations of real faces. Furthermore, Effort~\cite{effort} decomposes the feature space of CLIP~\cite{clip} into orthogonal subspaces, preserving pre-trained knowledge while learning forgery cues. 
However, these visual-only approaches overlook the audio modality that provides complementary cues for identifying audio-visual inconsistencies.

\noindent\textbf{Audio-Visual Deepfake Detection.}
Recent studies have increasingly explored leveraging both audio and visual cues for more reliable deepfake detection.
AVoid-DF \cite{avoid-df} detects multi-modal deepfakes by exploiting audio-visual inconsistency. 
AVGraph \cite{AVgraph} constructs heterogeneous audio-visual graphs to achieve fine-grained forgery classification.
PIA \cite{pia} incorporates language, facial motion, and identity cues, enabling a more comprehensive analysis. Several audio-visual methods also build upon pre-trained models. The authors of AVFF~\cite{oorloff2024avff} first pre-train the model to learn audio-visual correspondences in a self-supervised stage, followed by supervised deepfake classification.
AVPrompt~\cite{avprompt} fine-tunes CLIP~\cite{clip} and Whisper~\cite{whisper} for deepfake detection via prompt learning.
In addition, unsupervised methods, including AVAD~\cite{avad}, SpeechForensics~\cite{speechforensics}, and AVH-Align~\cite{avh-align}, determine authenticity by evaluating the matching degree between audio and video, without relying on explicit labels. However, the abundance of multi-modal information has, to some extent, diverted attention away from a deeper understanding of modality-specific information. 
\section{Method}
\label{sec:method}

\subsection{Overview}
\begin{figure}
    \centering
    \hspace*{-0cm} 
    \includegraphics[width=0.975\linewidth]{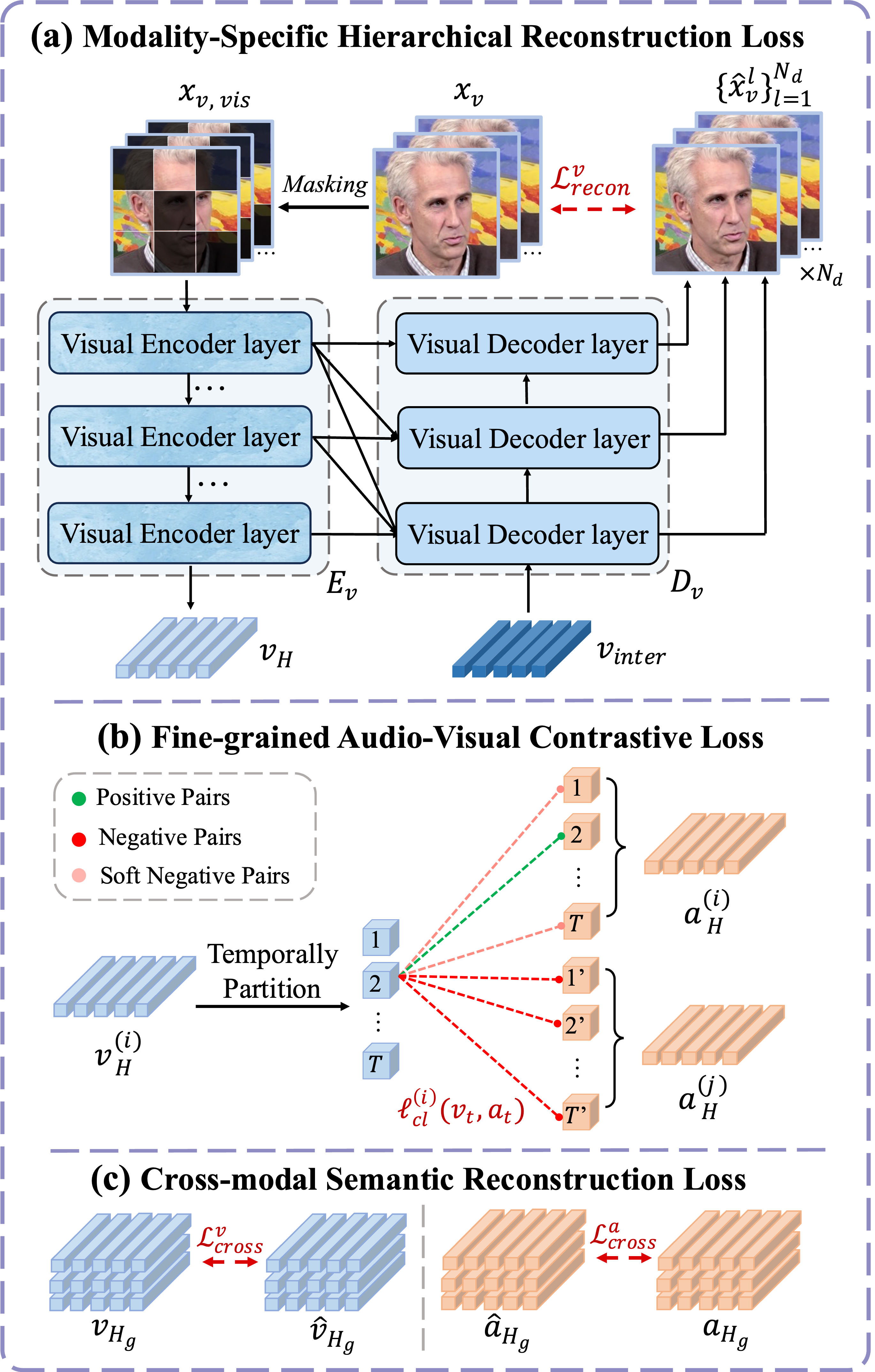}
    \vspace{-0.5em}
    \caption{\textbf{Illustrations of three self-supervised objectives in the Holistic Coherence Priors Pre-training phase}.
    (a) Each decoder layer reconstructs inputs using hierarchical encoder features, enforcing modality-specific structural coherence (illustrated with the visual modality).  
    (b) Audio and visual features are temporally partitioned and aligned segment by segment, with a soft negative pairs strategy to capture inter-modal micro-coherence (only one temporal segment and the visual-to-audio direction are illustrated for simplicity).  
    (c) One modality reconstructs the global semantic representation of the other, ensuring inter-modal macro-coherence.}
    \label{fig:3loss}
    \vspace{-2em}
\end{figure}

HAVIC is an audio-visual deepfake detector designed to models holistic coherence within and across modalities.
The training process for HAVIC is comprised of two stages. 
Firstly, we conduct self-supervised pre-training on a large-scale dataset of authentic videos to learn holistic coherence priors. 
Secondly, we fine-tune the model on a labeled dataset, enabling it to recognize deviations from authentic audio-visual features for deepfake detection.


Formally, given a video clip \(x_v \in \mathbb{R}^{t_v \times h \times w \times 3}\) (where \(t_v\) is the number of frames) with its accompanying audio, we first convert the audio waveform to a log-Mel spectrogram \(x_a \in \mathbb{R}^{t_a \times L}\) ($t_a$ and $L$ denote the number of audio frames and mel-frequency bins, respectively). 
We then adopt standard patch embeddings by dividing video frames into non-overlapping 3D spatio-temporal patches and the log-Mel spectrogram into 2D patches, then flattening and linearly projecting each patch to form $\bm{x}_v \in \mathbb{R}^{N_v \times C}$ and $\bm{x}_a \in \mathbb{R}^{N_a \times C}$, where $N_v$ and $N_a$ are the numbers of tokens and $C$ is the embedding dimension. 
We use two Transformer encoders, a visual encoder $E_v$ and an audio encoder $E_a$, to map the token sequences to hierarchical features.
Then we use an Audio-Visual Interaction Module $\mathcal{I}$ for cross-modal feature fusion. 
And finally, the model adaptively aggregate the hierarchical and interaction-aware features to produce the prediction.

In the following subsections, we elaborate on the training objectives and key designs adopted in the pre-training~(\ref{subsec:pretrain}) and fine-tuning~(\ref{subsec:finetune}) phases.

\subsection{Holistic Coherence Priors Pre-training}
\label{subsec:pretrain}

In pre-training, we adapt the MAE~\cite{MAE} strategy for each modality.
As shown in Figure~\ref{fig:pipeline_rl} (a), a random subset of tokens is masked, producing the masked tokens $\bm{x}_{v,{mask}}$, $\bm{x}_{a,mask}$ and the visible tokens $\bm{x}_{v,{vis}}$, $\bm{x}_{a,vis}$.
Visible tokens are fed into the encoders \(E_v\) and \(E_a\), and features are extracted from multiple intermediate layers and the final layer to form hierarchical representations \(\{\bm{v}_l\}_{l=1}^{H}\) and \(\{\bm{a}_l\}_{l=1}^{H}\), where $H$ is the total number of feature layer.

The high-level semantic features \(\bm{v}_H\) and \(\bm{a}_H\) are then passed to the Audio-Visual Interaction Module \(\mathcal{I}\), where bidirectional cross-attention integrates information from each modality into the other, yielding the interaction-aware features $\bm{v}_{inter}$ and $\bm{a}_{inter}$. 
These interaction-aware features are subsequently fed into the modality-specific decoders, $D_v$ and $D_a$, which have $N_d=H$ layers. Each decoding layer receives its corresponding encoder feature and shallower features to reconstruct the input. 
Meanwhile, the interaction-aware features are passed to the cross-modal semantic decoders, $D_{v \to a}$ and $D_{a \to v}$, which produce the reconstructed global semantic features of the counterpart modality, denoted as $\hat{\bm{a}}_{H_g}$ and $\hat{\bm{v}}_{H_g}$. The reconstruction targets, $\bm{a}_{H_g}$ and $\bm{v}_{H_g}$, are obtained by feeding all tokens into the encoders without gradient flow.

As shown in Figure~\ref{fig:3loss}, we design three synergistic objectives to learn holistic coherence in a self-supervised manner:
(1) a Modality-Specific Hierarchical Reconstruction Loss $\mathcal{L}_{rec}$ that learns comprehensive structural coherence,
(2) a Fine-grained Audio-Visual Contrastive Loss $\mathcal{L}_{cl}$ that captures temporal alignment and ensures inter-modal micro-coherence, and
(3) a Cross-modal Semantic Reconstruction Loss $\mathcal{L}_{cross}$ that enables each modality to reconstruct the global semantic representations of the other modality to achieve inter-modal macro-coherence.

\vspace{.4em}\noindent\textbf{Modality-Specific Hierarchical Reconstruction Loss.}
To learn the modality-specific coherence of each modality, we train the model to reconstruct the full input from a sparse subset of visible tokens. We enhance the learned representations with a hierarchical decoding design. 
Specifically, as illustrated in Figure~\ref{fig:3loss} (a), we use a stack of decoder layers, with each layer individually supervised to perform the full reconstruction task.
This process begins with the first decoder layer receiving the interaction-aware features ${\bm{v}_{inter}}$.
Its output is then passed to the next layer, and so on. 
Within each layer, the input feature first attends to the corresponding and shallower encoder features via cross-attention, then refined by a Transformer block, and finally projected to reconstruct the masked tokens.
The hierarchical reconstruction loss for the modality $m \in \{a,v\}$ is defined as:
\begin{equation}
\small
\mathcal{L}^{m}_{rec} = 
\frac{1}{N_d}\sum_{l=1}^{N_d} 
\left\| 
\hat{\bm{x}}^{l}_{m,mask} - \bm{x}_{m,mask}
\right\|_2^2,
\end{equation}
and the overall hierarchical reconstruction loss $\mathcal{L}_{rec}$ is obtained by summing over audio and visual modalities.



\vspace{.4em}\noindent\textbf{Fine-grained Audio-Visual Contrastive Loss.}
To ensure micro-coherence between the high-level semantic features $\bm{a}_H$ and $\bm{v}_H$ across different modalities of the same sample, we propose a fine-grained audio-visual contrastive learning strategy. 
As shown in Figure~\ref{fig:3loss} (b), we first partition $\bm{a}_H$ and $\bm{v}_H$ into $T$ temporal segments, denoted as $\{\bm{a}_{H,t}\}_{t=1}^T$ and $\{\bm{v}_{H,t}\}_{t=1}^T$. 
Audio and visual features from the same segment of the same sample are regarded as positive pairs, while all other pairs are treated as negatives.

To encourage the model to distinguish different segments within the same sample without over-penalizing their inherent similarity, we introduce a soft negative mechanism that adaptively adjusts the contribution of intra-sample negative pairs according to their temporal distance. the soft negative pairs weight $w_{t,t'}^{i,j}$ is defined as: \looseness=-1
\begin{equation}
\footnotesize
w_{t,t'}^{i,j} =
\begin{cases}
1 - 2 \cdot \sigma\!\left(-|t - t'|\right), & \text{if } i = j \text{ and } t \neq t', \\[4pt]
1, & \text{otherwise,}
\end{cases}
\end{equation}
where $i$ and $j$ are sample indices, $t$ and $t'$ denote the temporal segment indices of samples $i$ and $j$, and $\sigma(\cdot)$ is the sigmoid function. Formally, the fine-grained contrastive learning loss from the visual to the audio modality is defined as: \looseness=-1
\begin{equation}
{\scriptsize
\begin{aligned}
\mathcal{L}_{cl}(v,a) 
&= \frac{1}{2BT} \sum_{i=1}^B \sum_{t=1}^T 
\Big( - \log \ell_{cl}^{(i)} (v_t, a_t) \Big), \\[1pt]
\ell_{cl}^{(i)} (v_t, a_t) 
&= \frac{\exp(\text{sim}({\bar{v}}^{(i)}_t, {\bar{a}}^{(i)}_{t}) / \tau)}
{\sum_{j=1}^B \sum_{t'=1}^T  
\exp(\text{sim}({\bar{v}}^{(i)}_t, {\bar{a}}^{(j)}_{t'}) / \tau)\cdot w_{t,t'}^{i,j} },
\end{aligned}
}
\end{equation}
where $B$ is the batch size, 
$\text{sim}(\cdot, \cdot)$ is the cosine similarity, $\tau$ is the temperature parameter, and $\bar{m}^{(i)}_t$ denotes the mean semantic feature of the $t$-th segment from modality $m$ in the $i$-th sample. This loss is also computed in the reverse direction and summed to obtain the overall loss.

\vspace{.4em}\noindent\textbf{Cross-modal Semantic Reconstruction Loss.}
To further enhance macro-coherence between audio and visual modalities, we let the model reconstruct the global semantic features from the counterpart modality. As shown in Figure~\ref{fig:3loss} (c), the audio and video inputs (without masking) are fed into the audio and visual encoders to obtain the global semantic features $\bm{a}_{H_{g}}$ and $\bm{v}_{H_{g}}$, which serve as precise self-supervised targets for cross-modal semantic reconstruction. Gradients are stopped for $\bm{a}_{H_{g}}$ and $\bm{v}_{H_{g}}$, allowing them to act as fixed targets with negligible computational overhead. \looseness=-1

Specifically, we pad the interacted features $\bm{a}_{{inter}}$ and $\bm{v}_{{inter}}$ with learnable mask tokens and positional embeddings to reconstruct the complete structure, which is then fed into the corresponding cross-modal semantic decoders $D_{a\to v}$ and $D_{v\to a}$. The cross-modal semantic decoders first apply a linear projection layer to align the tokens with the number of tokens in the opposite modality. Then, a transformer block is applied for semantic refinement. Formally, the cross-modal semantic reconstruction loss for modality $m \in \{a,v\}$ is defined as:\looseness=-1
\begin{equation}
\small
\mathcal{L}^{m}_{{cross}} 
    = \left\| \bm{m}_{H_{{g}}} -\hat{\bm{m}}_{H_{{g}}} \right\|_2^2,
\end{equation}
where $\bm{m}_{H_g}$ is the global semantic feature, and $\hat{\bm{m}}_{H_g}$ is the reconstructed global semantic feature. The overall loss is obtained by summing over both audio and visual modalities.



The overall pre-training loss is defined as Eq.~\ref{eq:allloss}, where the $\lambda_*$ parameters represent the corresponding loss weights. 

\begin{equation}
    \label{eq:allloss}
    \mathcal{L}_{pt} = 
    \mathcal{L}_{rec} +
    \lambda_{cl}\,\mathcal{L}_{cl} + \lambda_{cross}\,\mathcal{L}_{cross}
\end{equation}


\subsection{Holistic Adaptive Aggregation Classification}
\label{subsec:finetune}

In the fine-tuning stage, we discard the decoders and feed the complete token sequences \(\bm{x}_v\) and \(\bm{x}_a\) into the visual and audio encoders \(E_v\) and \(E_a\).
Leveraging the learned holistic coherence priors, the encoders produce hierarchical representations \(\{\bm{v}_l\}_{l=1}^{H}\) and \(\{\bm{a}_l\}_{l=1}^{H}\) that span fine-grained local cues through to high-level semantic information. \looseness=-1


Furthermore, the pre-trained audio-visual interaction module \(\mathcal{I}\) is employed to extract interaction-aware features $\bm{v}_{inter}$ and $\bm{a}_{inter}$, thereby enhancing the capture of audio-visual correspondence or discrepancy.

The definitive evidence of a forgery may lie in low-level details, high-level semantics, or cross-modal discrepancies, and its nature can vary significantly between samples. 
Therefore, a mechanism to dynamically assess the importance of these different information sources is crucial. 
To this end, we introduce the Adaptive Feature Aggregation module $\mathcal{A}$, which learns to weigh and combine all available features to give the most effective evidence for the final prediction.
Taking the visual modality as an example, given the visual feature $\bm{v}_l \in \mathbb{R}^{N_v \times C}$ from \(\{\bm{v}_l\}_{l=1}^H\), a learnable scoring network $S_l(\cdot)$, implemented as a two-layer MLP with layers mapping from the embedding dimension to 128 and 1, is employed to estimate token-level importance scores of each token $\{t_{l_i}\}_{i=1}^{N_v}$. These scores are normalized by a softmax function to obtain importance weights, then the aggregated feature is computed as the weighted sum of tokens:

\begin{equation}
\bm{v}_{l,agg} = \sum_{i=1}^{N_v} \frac{\exp\big(S_l({t}_{l_i})\big)}{\sum_{j=1}^{N_v} \exp\big(S_l({t}_{l_j})\big)} {t}_{l_i}.
\end{equation}

Since tokens at different layers encode distinct structural and semantic information, we employ separate scoring networks $S_l(\cdot)$ for each layer, allowing for tailored importance evaluation according to the unique characteristics at each hierarchical level. 
Finally, we use learnable weights $\alpha$ to combine the aggregated features from different layers:
\begin{equation}
\bm{v}_{agg} = \sum_{l=1}^{H} \alpha_l \, \bm{v}_{l,agg},
\end{equation}
where \(\{\alpha_l\}_{l=1}^H\) are learnable weights that satisfy the normalization constraint \(\sum_{l=1}^H \alpha_l = 1\) and \(\alpha_l > 0\). 

The same procedure applies to the audio modality, producing \(\bm{a}_{agg}\).
The interaction-aware features \(\bm{v}_{inter}\) and \(\bm{a}_{inter}\) are also compressed using separate scoring networks to produce \(\bm{v}_{inter,agg}\) and \(\bm{a}_{inter,agg}\), respectively.

Then we employ a three-layer MLP as the main classifier, which takes \(\bm{a}_{agg}\), \(\bm{v}_{agg}\), \(\bm{a}_{inter,agg}\), and \(\bm{v}_{inter,agg}\) as input to predict whether a given sample is real or fake. 
To encourage the model to leverage information from both modalities and avoid over-reliance on the easier-to-detect modality, we introduce two auxiliary classifiers that operate on the uni-modal features \(\bm{a}_{agg}\) and \(\bm{v}_{agg}\), respectively.

All classifiers are trained using the standard cross-entropy loss, while only the main classifier is used during inference.
We detail the structure of the HAA Classification in the supplementary material.



\section{HiFi-AVDF Dataset}
\label{sec:dataset}
The traditional deepfake detection datasets \cite{faceforensics, dfdc, fakeavceleb, kodf} have significantly advanced research in deepfake detection.
However, most of their fake samples are generated by early GAN-based or face-swapping techniques \cite{GAN,fsgan,faceswap,faceshifter}, whose quality is relatively limited in comparison with recent generation models.
Therefore, we construct a new audio-visual deepfake benchmark dataset synthesized using multiple state-of-the-art generation models, namely the High-Fidelity Audio-Visual DeepFake (HiFi-AVDF) dataset. Our goal is to evaluate the generalization of existing detection models against these high-fidelity deepfakes.\looseness=-1

\begin{figure}[t]
    \centering
    \includegraphics[width=\linewidth]{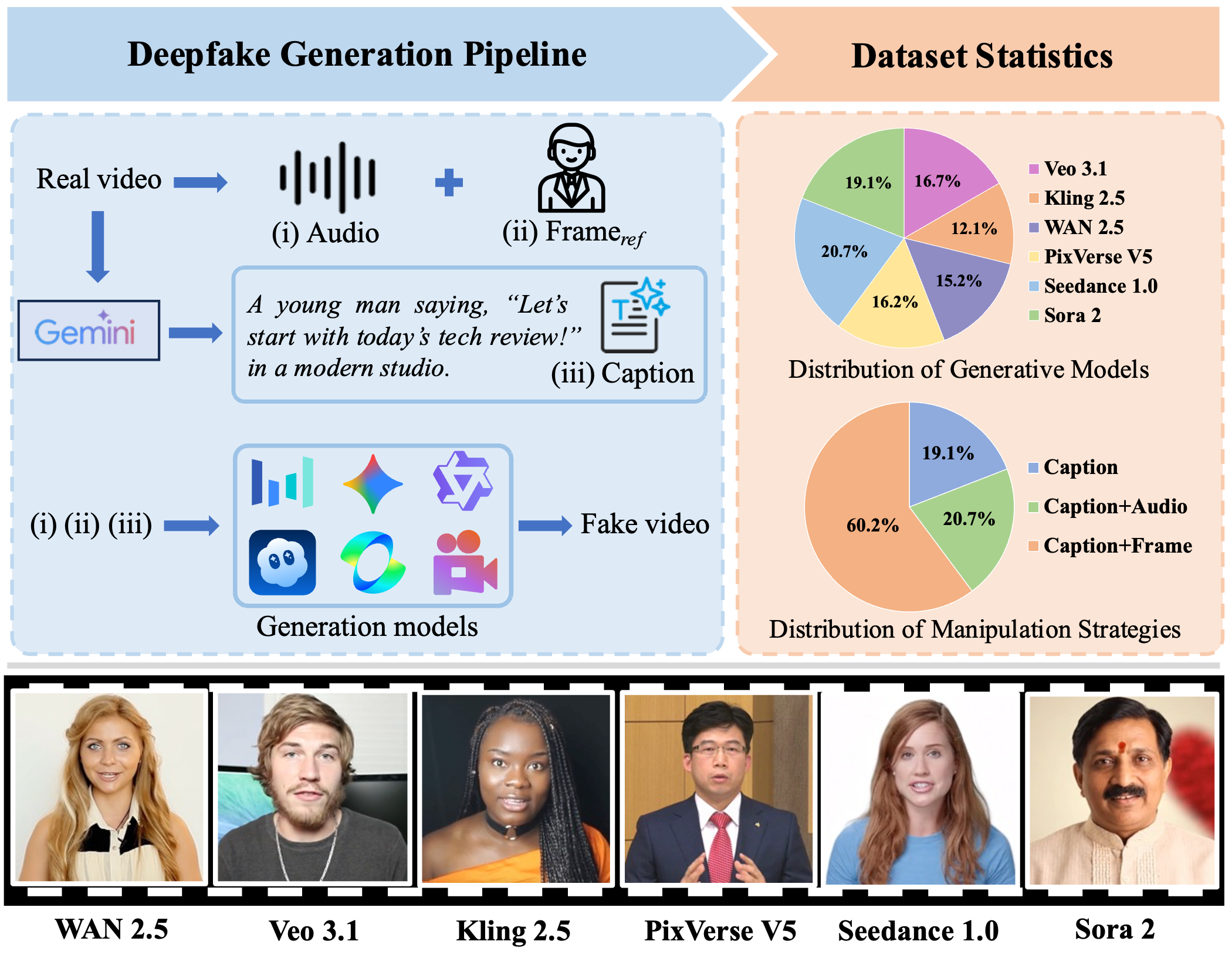}
    \vspace{-1.8em}  
    \caption{\textbf{Overview of the HiFi-AVDF dataset}. The left panel shows the deepfake generation process, where audio tracks, reference frames, and video captions of real videos are fed into generation models. The right panel shows dataset statistics, detailing the proportion of samples generated by each generative model and the distribution of different manipulation strategies. The bottom panels present representative examples generated by different models.}
    \label{fig:dataset}
\vspace{-1.0em}  
\end{figure}

\vspace{.4em}\noindent\textbf{Deepfake Generation Pipeline.}
We collect real videos from the AVSpeech dataset~\cite{AVSpeech}, which contains large-scale audio-visual clips originally sourced from YouTube, and perform a careful manual curation to ensure video quality. 
We exclude videos with poor visual quality, obvious background noise, and excessively short durations. 
After this screening process, approximately 2,000 high-quality videos remained. 
As shown on the left side of Figure~\ref{fig:dataset}, for each curated real video, we extract three core components: (1) a reference frame sampled from the video, (2) the audio track, and (3) a video caption generated by Gemini\footnote{Gemini API version in use: gemini-2.5-flash.}, which serve as inputs for the generators. 
We employ six state-of-the-art generation models: \textit{Kling 2.5}~\cite{kling2.5}, \textit{Veo 3.1}~\cite{veo3.1}, \textit{WAN 2.5}~\cite{wan2.5}, \textit{Seedance 1.0}~\cite{seedance}, \textit{PixVerse V5}~\cite{pixverse}, and \textit{Sora 2}~\cite{sora2}, to generate corresponding forgeries for each real video.
We detail the generation pipeline in the supplementary material.

\vspace{.4em}\noindent\textbf{Dataset Statistics.}
The HiFi-AVDF dataset comprises a total of 3,810 high-quality videos, including 1,905  generated videos paired with corresponding real videos, all containing both visual and audio tracks. To mitigate potential bias due to differences in frame rate, both real and generated videos are captured at 25 fps, with durations ranging from 4 to 10 seconds. 
The dataset statistics are displayed on the right, and representative examples generated by different models are illustrated at the bottom of Figure~\ref{fig:dataset}.
Further statistics on HiFi-AVDF and a more comprehensive comparison with related works are provided in the supplementary material.

\section{Experiments}
\label{sec:experiments}
\subsection{Experimental Setup}
\noindent\textbf{Datasets and metrics.}
We first pre-train the HAVIC on the LRS2 dataset~\cite{lrs}, which contains only real videos, to learn coherence priors.
We then fine-tune the model on the FakeAVCeleb dataset~\cite{fakeavceleb}.
After training, we evaluate on three benchmarks: FakeAVCeleb for intra-dataset performance, and both our proposed HiFi-AVDF and KoDF \cite{kodf} for cross-dataset generalization. Additional details on the datasets are provided in the supplementary material.


We evaluate performance using accuracy (ACC), average precision (AP), and area under the ROC curve (AUC), each averaged over multiple runs with different random seeds to mitigate randomness and ensure reliable comparison.

\begin{table}[t]
    \centering
    \renewcommand{\arraystretch}{1.02}
    {\fontsize{8}{9}\selectfont
    \begin{adjustbox}{width=0.9\linewidth,center}
    \begin{tabular}{r c *{2}{c}}
    \toprule
    Method & Modality & ACC & AUC   \\
    \midrule
    Xception \cite{Xception} & V  & 67.9 & 70.5  \\
    LipForensics \cite{lip-forensics} & V & 80.1 & 82.4  \\
    FTCN \cite{ftcn} & V  & 64.9 & 84.0\\
    RealForensics \cite{realforensics} & V & 89.9 & 94.6\\
    \midrule
    AVoiD-DF \cite{avoid-df} & AV & {83.7} & {89.2}  \\
    MRDF-CE \cite{mrdf} & AV & 94.1 & 92.4\\
    AVFF \cite{oorloff2024avff} & AV & 98.6 & 99.1 \\
    PIA \cite{pia} & AV & \underline{98.7} & \underline{99.8} \\
    \midrule
    \textbf{HAVIC (Ours)} & \textbf{AV} & \textbf{99.8} & \textbf{99.9} \\
    \bottomrule
    \end{tabular}
    \end{adjustbox}
    }
    \vspace{-0.5em}
    \caption{\textbf{Intra-Dataset Performance on FakeAVCeleb.} 
    Best result is in bold, and second best is underlined.
    }
    \vspace{-2em}
    \label{tab:intra_eval}
\end{table}
 
\vspace{.4em}\noindent\textbf{Implementation details.}
Following \cite{oorloff2024avff}, we sample video clips of 3.2s in duration, from which visual frames and corresponding audio waveforms are extracted at 5 fps and 16 kHz, respectively. Facial regions are cropped from visual frames using FaceX-Zoo \cite{facexzoo} to eliminate background interference and are spatially resized to 224 × 224. The audio is transformed into a Mel-spectrogram with 128 Mel-frequency bins and 1024 time frames.
We use $N_e=12$ layers in encoders, $N_d=4$ layers in decoders, and $H=4$ layers in hierarchical features, which are uniformly extracted from the 3rd, 6th, 9th, and 12th layers of the encoders. 
The number of temporal segments in the Fine-grained Audio-Visual Contrastive Loss is set to $T=8$.
Please refer to the supplementary for additional details on implementation.

\subsection{Main Results}
\noindent\textbf{Intra-Dataset Performance.}
For fair comparisons, we follow the same data split protocol in~\cite{avoid-df, oorloff2024avff, pia}, using 70\% of the FakeAVCeleb samples for training and validation, and the remaining 30\% for testing. We report ACC and AUC as evaluation metrics, consistent with prior works~\cite{avoid-df, oorloff2024avff, pia} to ensure comparability. As shown in Table~\ref{tab:intra_eval}, our method achieves the best performance among all competitors, reaching 99.8\% in ACC and 99.9\% in AUC. 
Compared to visual-only approaches~\cite{Xception,ftcn,lip-forensics,realforensics}, 
our method substantially mitigates the limitations of single-modality detection by effectively leveraging complementary cues from both audio and video.
In contrast to previous multi-modal baselines~\cite{avoid-df,mrdf,oorloff2024avff},
our approach further improves performance by not only modeling the audio-visual correspondence but also explicitly enforcing modality-specific coherence within each modality.
Although PIA~\cite{pia} achieves high performance by incorporating language, face motion, and facial identification cues, our method attains even stronger results without relying on auxiliary sources, demonstrating its effectiveness in capturing comprehensive coherence in a more streamlined way. 

\vspace{.4em}\noindent\textbf{Cross-Dataset Generalization on KoDF.}
To examine the cross-dataset generalization capability, we evaluate the model trained on FakeAVCeleb~\cite{fakeavceleb} using a subset of KoDF~\cite{kodf}, following the settings in \cite{avad, oorloff2024avff}. We report AP and AUC as generalization evaluation metrics, consistent with prior works~\cite{avad, oorloff2024avff} to ensure comparability. As shown in Table~\ref{tab:kodf}, our method demonstrates superior cross-dataset generalization on KoDF, outperforming all existing audio-visual and visual-only baselines, achieving 99.2\% AP and 98.9\% AUC.
The remarkable cross-dataset performance of AVFF~\cite{oorloff2024avff}, RealForensics~\cite{realforensics}, and our method can be attributed to pre-training on real videos. 
RealForensics~\cite{realforensics} learn temporally dense video representations from real videos to capture facial cues, 
while AVFF~\cite{oorloff2024avff} achieves high generalization by aligning real audio and video features to capture cross-modal consistency.
Compared to RealForensics~\cite{realforensics}, our method achieves an increase in AP of 3.5\% (+5.3\% in AUC),
and compared to AVFF~\cite{oorloff2024avff}, our method achieves an increase in AP of 6.1\% (+3.4\% in AUC), 
highlighting the effectiveness of our holistic audio-visual coherence modeling in enhancing generalization.

\begin{table}[b]
    \centering
    \renewcommand{\arraystretch}{1.02}
    {\fontsize{8}{9}\selectfont
    \begin{adjustbox}{width=0.9\linewidth,center}
    \begin{tabular}{r c *{2}{c} }
    \toprule
    Method & Modality & AP & AUC \\
    \midrule
    Xception \cite{Xception} &V & 76.9 & 77.7\\
    LipForensics \cite{lip-forensics} &V & 89.5 & 86.6\\  
    FTCN \cite{ftcn} &V & 66.8 & 68.1\\ 
    RealForensics~\cite{realforensics}  &V & \underline{95.7} & 93.6\\ 
     \midrule
    *AVAD \cite{avad}  & AV & 87.6 & 86.9 \\
    *AVH-Align \cite{avh-align}  & AV & 88.3 & 90.1 \\
    AVFF \cite{oorloff2024avff} & AV & 93.1 & \underline{95.5}\\
    PIA \cite{pia} & AV & 91.7 & 95.0 \\
    
    \midrule
    \textbf{HAVIC (Ours)} & \textbf{AV} & \textbf{99.2} & \textbf{98.9}\\
    \bottomrule
    \end{tabular}
    \end{adjustbox}
    }
    \vspace{-0.5em}
    \caption{\textbf{Cross-Dataset Generalization on KoDF.}  All supervised methods are trained on FakeAVCeleb for fair comparison.
    Methods with * denote unsupervised ones. 
    }
    \label{tab:kodf}
\end{table}

\begin{table*}[t]
    \centering
    \renewcommand{\arraystretch}{1.3}
    \begin{adjustbox}{width=0.95\linewidth,center}
    \begin{tabular}{r c *{12}{c} *{2}{c}}
    \toprule
    \multirow{2}{*}{Method} & \multicolumn{2}{c}{Veo 3.1} & \multicolumn{2}{c}{Kling 2.5} & \multicolumn{2}{c}{Seedance 1.0} & \multicolumn{2}{c}{PixVerse V5} & \multicolumn{2}{c}{WAN 2.5} & \multicolumn{2}{c}{Sora 2}  & \multicolumn{2}{c}{AVG} \\
    \cmidrule(lr){2-3} \cmidrule(lr){4-5} \cmidrule(lr){6-7} \cmidrule(lr){8-9} \cmidrule(lr){10-11} \cmidrule(lr){12-13} \cmidrule(lr){14-15}
     & AP & AUC & AP & AUC & AP & AUC & AP & AUC & AP & AUC & AP & AUC & AP & AUC \\
    \midrule
    *AVAD~\cite{avad} & 62.23 & \textbf{67.07} & 60.72 & 62.09 & 56.72 & 59.26 & 59.54 & 59.05 & 61.60 & 61.12 & 59.39 & 61.87 & 60.37 & 61.73 \\
    *AVH-Align~\cite{avh-align} & 63.84 & 56.76 & 59.11 & 55.71 & 51.27 & 49.04 & \underline{74.87} & \textbf{75.84} & 59.73 & 57.36 & 57.28 & 59.23 & 61.35 & 58.32 \\
    RealForensics~\cite{realforensics} & 61.25 & 58.17 & 61.48 & 62.93 & 64.38 & 65.72 & 57.44 & 60.51 & 61.74 & 64.85 & 60.43 & 59.25 & 61.12 & 61.91 \\
    LipFD~\cite{lipfd} & 56.43 & 59.52 & 62.49 & 68.51 & 61.43 & 67.62 & 57.21 & 60.23 & 54.25 & 61.85 & 65.78 & 56.11 & 59.60 & 62.31 \\
    AVFF~\cite{oorloff2024avff} & 63.92 & 60.60 & \underline{81.12} & \underline{78.27} & \underline{70.36} & \underline{74.43} & 59.15 & 53.86 & \underline{67.08} & \underline{74.38} & 54.46 & 50.15 & \underline{66.02} & 65.28 \\
    Effort~\cite{effort} & \textbf{70.62} & 62.04 & 63.76 & 70.64 & 57.62 & 58.98 & 62.78 & 59.34 & 66.57 & 66.94 & \textbf{68.49} & 67.26 & 64.97 & 64.20 \\
    PIA~\cite{pia} & 58.87 & 65.09 & 66.42 & 73.08 & 64.30 & 64.97 & 53.73 & 57.15 & 59.24 & 65.17 & 65.16 & \textbf{72.91} & 61.29 & \underline{66.40} \\
    \midrule
    \textbf{HAVIC (Ours)} & \underline{68.74} & \underline{65.92} & \textbf{85.53} & \textbf{84.72} & \textbf{76.81} & \textbf{81.07} & \textbf{77.58} & \underline{74.39} & \textbf{76.47} & \textbf{78.02} & \underline{67.35} & \underline{70.52} & \textbf{75.41} & \textbf{75.77} \\
    \bottomrule
    \end{tabular}
    \end{adjustbox}
    \vspace{-0.5em}
    \caption{\textbf{Cross-Dataset Generalization on HiFi-AVDF.}
    We evaluate models trained on traditional datasets against synthetic videos generated by state-of-the-art generation models. For fair comparison, all methods trained on the FakeAVCeleb \cite{fakeavceleb} dataset. 
    }
    \label{tab:hifi-avdf}
    \vspace{-1.5em}
\end{table*}

\vspace{.4em}\noindent\textbf{Cross-Dataset Generalization on HiFi-AVDF.}
We further evaluate the generalization ability of the model by testing it on our proposed HiFi-AVDF dataset, and conduct comparisons with state-of-the-art multi-modal detectors.
Results are reported in Table~\ref{tab:hifi-avdf}.
Compared to the cross-dataset results on KoDF, all evaluated methods, including ours, exhibit a noticeable performance decline. This discrepancy arises because KoDF consists of relatively earlier-generation deepfakes, whereas HiFi-AVDF contains more photorealistic and tightly synchronized deepfakes generated by cutting-edge models.
Despite this difficulty, our HAVIC achieves the highest average AP (75.41\%) and AUC (75.77\%) across all generation models, significantly outperforming existing state-of-the-art methods with improvements of 9.39\% AP and 9.37\% AUC, demonstrating strong generalization to high-fidelity audio-visual deepfakes.

\subsection{Ablation Studies}
We conduct comprehensive ablation studies to examine the contribution of each component in our framework. For clarity, we present the core results in the main paper. Please refer to the supplementary material for 
the hyperparameter sensitivity experiment on $\lambda_{*}$, the audio-missing ablation, comparisons with other self-supervised learning methods, and analyses of model parameters and computational cost.

\noindent\textbf{Effectiveness of HCP Pre-training.}
We first assess the effectiveness of the Holistic Coherence Priors (HCP) pre-training, as shown in Table~\ref{tab:ablation_HCP}. Skipping this phase leads to a substantial performance drop across all datasets. Excluding $\mathcal{L}_{rec}$ forces the model to rely solely on modality-specific coherence prior, thereby impairing its ability to capture inter-modal coherence and lowering performance. Removing $\mathcal{L}_{cl}$ or $\mathcal{L}_{cross}$, which ensure inter-modal coherence prior, consistently degrades all metrics and highlights the necessity of learning both modality-specific and inter-modal coherence priors to improve deepfake detection performance.


\begin{table}[b] 
\vspace{-1em}
\centering \renewcommand{\arraystretch}{1.1}
\begin{adjustbox}{width=0.95\linewidth,center} \begin{tabular}{l cc cc cc} \toprule \multirow{2}{*}{Method} & \multicolumn{2}{c}{FakeAVCeleb} & \multicolumn{2}{c}{KoDF} & \multicolumn{2}{c}{HiFi-AVDF} \\ \cmidrule(lr){2-3} \cmidrule(lr){4-5} \cmidrule(lr){6-7} & ACC & AUC & AP & AUC & AP & AUC \\ \midrule  Ours w/o HCP Pre-training & 83.5 & 80.7 & 63.4 & 71.0 & 58.3 & 57.6 \\  Ours w/o $\mathcal{L}_{{rec}}$ & 91.4 & 94.5 & 87.4 & 90.2 & 67.3 & 65.6 \\ Ours w/o $\mathcal{L}_{{cl}}$ & 97.4 & 98.6 & 92.5 & 94.4 & 66.8 & 64.2 \\  Ours w/o $\mathcal{L}_{{cross}}$ & 98.7 & 99.2 & 96.7 & 97.2 & 73.2 & 72.3 \\ \midrule \textbf{HAVIC (Ours)} & \textbf{99.8} & \textbf{99.9} & \textbf{99.2} & \textbf{98.9} & \textbf{75.4} & \textbf{75.7} \\ \bottomrule \end{tabular} \end{adjustbox} \vspace{-0.5em} \caption{Ablations study on HCP Pre-training.} \label{tab:ablation_HCP} \vspace{-1.5em} \end{table}

\noindent\textbf{Contribution of HAA Classification.}
Table~\ref{tab:HAA_ablation} presents the contribution of the Hierarchical Adaptive Aggregation (HAA) Classification.
Without the Adaptive Aggregation and using mean pooling instead, the model can only average features uniformly, failing to dynamically emphasize discriminative cues.
Removing the auxiliary classifiers, the model lacks sufficient modality-specific supervision.
Each component brings partial gains, while their combination yields the best overall performance. 
\begin{table}[h]
\centering
\vspace{-0.5em}
\renewcommand{\arraystretch}{1.1}
\begin{adjustbox}{width=0.95\linewidth,center}
\begin{tabular}{cccccccc}
\toprule
\multirow{2}{*}{\makecell{Adaptive \\ Aggregation}} & \multirow{2}{*}{\makecell{Auxiliary \\ Classifiers}} & \multicolumn{2}{c}{FakeAVCeleb} & \multicolumn{2}{c}{KoDF} & \multicolumn{2}{c}{HiFi-AVDF} \\ 
\cmidrule(lr){3-4} \cmidrule(lr){5-6} \cmidrule(lr){7-8}
 & & ACC & AUC & AP & AUC & AP & AUC \\
\midrule
\textcolor{red}{\xmark} & \textcolor{red}{\xmark} & 98.8 & 99.0 & 94.8 & 93.4 & 65.3 & 67.6 \\
\textcolor{green}{\cmark} & \textcolor{red}{\xmark} & 99.5 & 99.3 & 95.9 & 95.6 & 67.7 & 71.8 \\
\textcolor{red}{\xmark} & \textcolor{green}{\cmark} & 99.4 & 99.5 & 97.8 & 95.9 & 71.2 & 72.9 \\
\midrule
\textcolor{green}{\cmark} & \textcolor{green}{\cmark} & \textbf{99.8} & \textbf{99.9} & \textbf{99.2} & \textbf{98.9} & \textbf{75.4} & \textbf{75.7} \\
\bottomrule
\end{tabular}
\end{adjustbox}
\vspace{-0.5em}
\caption{Ablation study on HAA Classification, where \textcolor{red}{\xmark} and \textcolor{green}{\cmark} denote the removal and inclusion of the component, respectively.}
\label{tab:HAA_ablation}
\vspace{-1.5em}
\end{table}

\section{Conclusion}
\label{sec:conclusion}
In this paper, we introduce HAVIC, a novel framework that leverages holistic coherence within and across modalities for deepfake detection. 
Furthermore, we present HiFi-AVDF, a high-fidelity audio-visual deepfake dataset designed to benchmark detection methods against state-of-the-art commercial generation models. 
Our extensive experiments demonstrate that HAVIC achieves superior performance across various datasets, excelling in both intra-dataset and cross-dataset generalization scenarios, and outperforming prior methods by 9.39\% AP and 9.37\% AUC on the challenging cross-dataset scenario.
\newline
\textbf{Limitations.}
Since the datasets we used primarily consist of face videos, our work is limited to human-face forgery cases. Transferring the model to non-human scenarios (e.g., forged animal, object, or scene videos) can be challenging.
\newline
\textbf{Future Works.} 
While HAVIC demonstrates strong generalization, the performance of all evaluated methods, including ours, significantly decreases on HiFi-AVDF dataset. This highlights the need for future research to develop deepfake detection techniques that are more resilient against manipulations from cutting-edge generation models, as well as robust to a wider range of potential adversarial attacks.

\clearpage
\setcounter{page}{1}
\maketitlesupplementary
\setcounter{section}{0}
\setcounter{figure}{0}
\setcounter{table}{0}
\renewcommand{\thesection}{\Alph{section}}


\section{Overview}
In this supplementary material, we provide additional details of the proposed HAVIC framework and the HiFi-AVDF dataset, as well as more experimental results.  \cref{supp_sec:implementation} introduces further implementation details, including model architecture choices and training configurations. \cref{supp_sec:datasets} then offers a comprehensive description of the proposed HiFi-AVDF dataset and information about other datasets. Finally, \cref{supp_sec:results} reports additional results, including extended evaluations, ablation studies, and further analyses.

\section{Implementation Details} \label{supp_sec:implementation}
\subsection{Holistic Coherence Priors Pre-training}
\noindent\textbf{Inputs.}
We sample 3.2s video clips from the LRS2 dataset~\cite{lrs} and preprocess them following the procedure described in Sec. 5.1 in the main paper. Note that the LRS2 dataset contains only real videos. Each clip is converted into 16 cropped face frames, which are then resized to 224$\times$224 to serve as the visual input.
For the audio stream, we extract 128-dimensional log Mel filterbank features using a 25 ms Hanning window and a 10 ms hop length. The resulting spectrogram is subsequently resized to 1024$\times$128 and used as the audio input.
Both audio and visual inputs are normalized and then tokenized. \looseness=-1

\noindent\textbf{Token Masking.}
In the Holistic Coherence Priors Pre-training phase, we adopt the MAE framework~\cite{MAE} to enable efficient self-supervised pre-training, in which a large proportion of tokens are masked. 
For visual tokens, we deploy tube masking~\cite{tong2022videomae} with a ratio of 90\%,
where patches at the same spatial location across each frame share the same mask to reduce temporal information leakage.
As for audio tokens, random masking~\cite{audiomae} with a ratio of 81.25\% is applied to achieve diverse time–frequency coverage. The masking ratios are chosen empirically based on previous research~\cite{tong2022videomae, audiomae}. \looseness=-1

\noindent\textbf{Model Architecture.}
We adopt a symmetric architecture for both audio and visual modalities.
Each encoder consists of $N_e=12$ Transformer layers, with 12 attention heads per layer and an embedding dimension of 768. 
Hierarchical features are extracted from $H=4$ layers, uniformly selected from the 3rd, 6th, 9th, and 12th layers of the encoders. 
The audio-visual interaction module comprises an 8-head cross-attention layer followed by an 8-head Transformer block.
The cross-modal semantic decoders include a linear projection layer followed by a single Transformer block with 12 attention heads.
The modality-specific decoders consist of $N_d=4$ Transformer layers with 6 attention heads per layer and an embedding dimension of 384, four 6-head cross-attention layers corresponding to the hierarchical feature layers, and four linear heads that project each layer's output back to the original input.

\noindent\textbf{Training Configuration.}
Following \cite{oorloff2024avff}, we initialize the audio encoder–decoder with pretrained AudioMAE \cite{audiomae} weights from AudioSet-2M \cite{gemmeke2017audio}, and initialize the visual encoder–decoder with MARLIN \cite{marlin} pretrained on the YouTubeFace dataset \cite{wolf2011face}.
The Fine-grained Contrastive loss weight is set empirically to $\lambda_{cl} = 0.01$, and the temperature parameter $\tau$ is fixed at 0.07. The Cross-modal Semantic Reconstruction loss weight is fixed at 1. 
Both $\mathcal{L}_{rec}$ and  $\mathcal{L}_{cross}$ are computed by averaging squared errors over the tokens.
Subsequently, we pre-train the HAVIC using the AdamW optimizer~\cite{loshchilov2018decoupled} with a learning rate of 1.5e-4 with a cosine decay~\cite{loshchilov2016sgdr}.
We train for 200 epochs with a linear warmup for 20 epochs using four NVIDIA L20 GPUs with a total batch size of 112. It takes about five days to complete the Holistic Coherence Priors Pre-training phase.

\begin{figure*}[t]
    \centering
    \includegraphics[width=1.\linewidth]{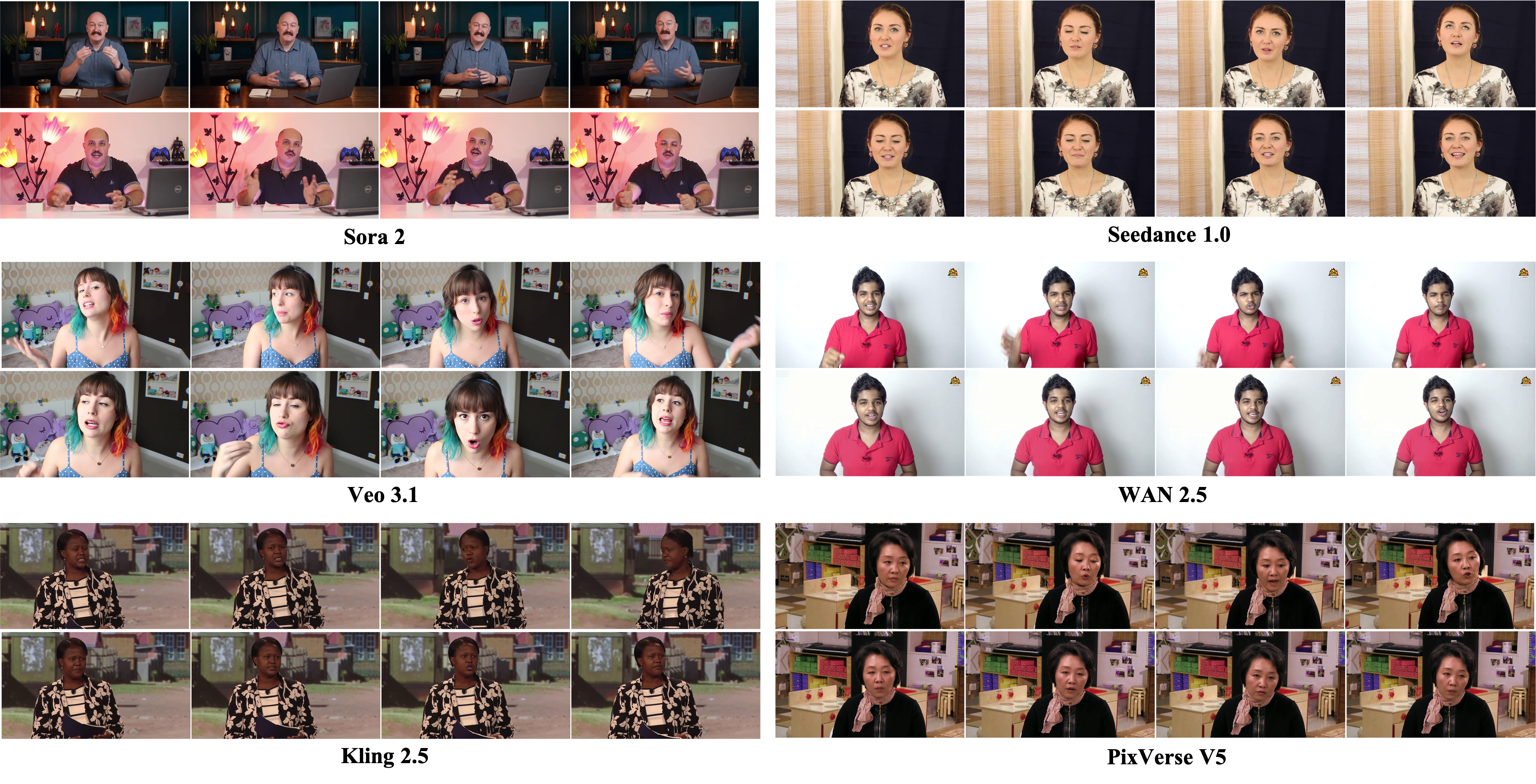}
    \vspace{-2em}
    \caption{\textbf{Examples of real–fake video pairs generated by the six models in HiFi-AVDF.}
For each model, we display one representative pair, where the top row shows a real video clip and the bottom row shows the corresponding forged clip produced by that model.}
    \label{fig:hifi-avdf-pairs}
        \vspace{-1em}
\end{figure*}

\subsection{Holistic Adaptive Aggregation Classification}
\noindent\textbf{Inputs.}
The inputs for the Holistic Adaptive Aggregation Classification phase are drawn from the FakeAVCeleb~\cite{fakeavceleb} dataset, which contains deepfake videos with manipulated audio, visual, or both modalities.
The processing procedure follows that of the Holistic Coherence Priors pre-training phase, with the difference that no masking is applied to the input in this phase.
Following~\cite{oorloff2024avff}, we apply weighted sampling to alleviate class imbalance between real and fake samples in the FakeAVCeleb dataset.

\noindent\textbf{Model Architecture.}
We remove the decoders of HAVIC. The audio and visual encoders, along with the audio-visual interaction module, retain the same structure as in the Holistic Coherence Priors pre-training phase. Each scoring network in the Adaptive Feature Aggregation module is a 2-layer MLP, and each classification head is a 3-layer MLP. 

\noindent\textbf{Training Configuration.}
We initialize the audio encoder, visual encoder, and audio-visual interaction module using the weights obtained from the Holistic Coherence Priors Pre-training phase. The model is then trained using the AdamW optimizer~\cite{loshchilov2018decoupled} with a cosine annealing scheduler with warm restarts~\cite{loshchilov2016sgdr}. A smaller learning rate of 1.0e-5 is applied to the pretrained components, while newly added modules are trained with a larger learning rate of 1.0e-4. Training is performed for 50 epochs with a total batch size of 32 on four NVIDIA L20 GPUs, taking about eight hours.

\noindent\textbf{Inference.}
During inference, we follow \cite{oorloff2024avff} and apply a sliding-window strategy for video-level detection. Each window has a duration of 3.2s and slides with a step size of 0.4s. The output logits from the main classification head are computed for each window, and the final prediction (real or fake) is obtained by averaging the logits across all windows.
\section{Dataset Details} \label{supp_sec:datasets}
\subsection{HiFi-AVDF Dataset}

\noindent\textbf{Overall Dataset Creation Pipeline.}
After collecting the real data and extracting three core components, namely the reference frame, audio track, and video caption, as described in Sec. 4 in the main paper, forged videos are generated following three strategies to simulate diverse audio-visual manipulations: \newline
(i) \textbf{caption-only}: Video generation is driven entirely by a text caption, creating a purely text-to-video (T2V) generation without audio-visual references. \newline
(ii) \textbf{caption + reference frame}: A reference frame is incorporated with the text caption to ground the generated content, significantly improving visual realism.\newline
(iii) \textbf{audio track + reference frame}: The generator synchronizes the lip movements of a subject in a reference frame with a source audio track, producing realistic, audio-visually coherent forgeries.

In practice, Sora 2 follows the caption-only strategy, as it does not support using a reference face image for video generation. Seedance 1.0 employs the audio track + reference frame strategy, while the remaining four models utilize the caption + reference frame strategy. As shown in Fig.~\ref{fig:hifi-avdf-pairs}, for each real video, a corresponding forged video is generated using one of the models, resulting in a dataset comprising 1,905 real videos and 1,905 corresponding fake videos. 

\noindent\textbf{Comparison with Existing Datasets.}
We compare HiFi-AVDF with representative deepfake datasets in terms of manipulated modality, dataset curation, availability of text-to-video (T2V) and image-to-video (I2V) samples, generation methods, number of person, and the counts of real and fake samples. As summarized in Tab.~\ref{tab:sup_dataset_summary}, HiFi-AVDF provides high-quality audio-visual forgeries generated using diverse state-of-the-art methods, covering a larger number of person and offering both T2V and I2V capabilities. This comprehensive design enables more robust evaluation of audio-visual deepfake detection models and facilitates research on multi-modal forgery scenarios.

\renewcommand{\arraystretch}{1.}
\begin{table*}[t]
\centering
\resizebox{\textwidth}{!}{
\begin{tabular}{lccccccccc}
\toprule
Dataset & \makecell{Manipulated \\ Modality} & T2V & I2V & Generation Method & Person \# & Real \# & Fake \# & Year \\
\midrule

FaceForensics++~\cite{Xception} & V & \xmark&\xmark& \makecell{FaceSwap (2017)~\cite{faceswap}, DeepFakes (2017)\cite{deepfakes}, \\ Face2Face (2016)\cite{face2face}, NeuralTextures (2019)\cite{neuraltextures}} & N/A & 1,000 & 4,000 & 2019  \\
WildDeepfake~\cite{wilddeepfake} & V &\xmark&\xmark &  N/A  & N/A & 3,805 & 3,509 & 2020 \\
KoDF~\cite{kodf} & V &\xmark &\xmark & \makecell{FaceSwap~\cite{faceswap} (2017), DeepFaceLab~\cite{deepfacelab}(2020),  FOMM~\cite{fomm} (2019) \\ FaceSwapGAN~\cite{fsgan} (2019), ATFHP~\cite{atfhp} (2020), Wav2Lip~\cite{wav2lip} (2020)} & 403 & 62,166 & 175,776 & 2021 \\
DF-Platter~\cite{df-platter} & V &\xmark&\xmark& FaceSwap (2017)~\cite{faceswap}, FaceSwapGAN (2019)~\cite{fsgan}, Faceshifter (2019)~\cite{faceshifter} & 454  & 133,260 & 132,496  & 2023\\
FakeAVCeleb~\cite{fakeavceleb} & AV &\xmark&\xmark& \makecell{Wav2Lip (2020)~\cite{wav2lip} , FaceSwap (2017)~\cite{faceswap},\\ FaceSwapGAN (2019)~\cite{fsgan}, SV2TTS (2018)~\cite{sv2tts}} & 500  & 500 & 19,500 & 2021 \\
DefakeAVMiT~\cite{avoid-df}  & AV &\xmark&\xmark& \makecell{FaceSwap (2017)~\cite{faceswap}, Voice Replay (2017)~\cite{voice-replay}, SV2TTS (2018)~\cite{sv2tts}, DeepFaceLab (2020)~\cite{deepfacelab}, \\ Wave2Lip (2020)~\cite{wav2lip}, PC-AVS (2021)~\cite{pc-avs},  EVP (2021)~\cite{evp}, AV exemplarAE (2020)~\cite{av-exemplar-autoencoders}}& 86 & 540 & 6,480 & 2023 \\
AV-Deepfake1M~\cite{avdeepfake1m} & AV &\xmark&\xmark& VITS (2021)~\cite{vits}, YourTTS (2022)~\cite{yourtts}, TalkLip (2023)~\cite{talklip} & 2068 & 286,721 & 860,039 & 2024 \\
\midrule
\textbf{HiFi-AVDF (ours)} & AV &\cmark &\cmark& \makecell[c]{Sora 2 (2025)~\cite{sora2}, Veo 3.1 (2025)~\cite{veo3.1}, Seedance 1.0 (2025)~\cite{seedance}, \\ Kling 2.5 (2025)~\cite{kling2.5}, WAN 2.5 (2025)~\cite{wan2.5}, PixVerse V5 (2025)~\cite{pixverse}} & 1905 & 1,905 & 1,905 & 2025  \\ 
\bottomrule
\end{tabular}
}
\caption{\textbf{Summary of representative deepfake datasets.} This table compares commonly used visual and audio-visual deepfake datasets in terms of manipulated modality, support for T2V/I2V generation, generation methods, number of person, and counts of samples. HiFi-AVDF provides high-quality audio-visual forgeries generated by diverse state-of-the-art models, supporting both T2V and I2V modalities.}
\label{tab:sup_dataset_summary}
\vspace{-1em}
\end{table*}

\noindent\textbf{Ethical and Bias Issues.} 
We acknowledge that the HiFi-AVDF dataset may raise ethical concerns, particularly regarding the potential misuse of facial videos and the advanced audio-visual generation tools employed in constructing the dataset. Such misuse could involve the creation of new deepfake content or other forms of malicious exploitation. To mitigate these risks, we release the dataset under a carefully designed end-user license agreement that explicitly restricts the use of the dataset and any generated audio-visual content to research purposes only. The dataset is provided solely to support scientific progress in deepfake detection, and any attempt to employ the data for harmful or non-research activities is strictly prohibited.

Furthermore, to conduct a comprehensive bias assessment of HiFi-AVDF, we perform an automated demographic analysis using EasyFace~\cite{easyface}, categorizing individuals by binary gender, seven racial/ethnic groups, and nine age ranges spanning from infancy to older adulthood. This procedure provides a fine-grained understanding of the dataset’s demographic composition. A face detection pipeline equipped with pre-trained multi-attribute recognition models was applied to all samples, and the resulting statistics are summarized in Tab.~\ref{tab:demographics}. 
While our dataset provides broad demographic coverage, it still contains certain degrees of demographic bias. Additionally, automated demographic classification may be inaccurate for edge cases and intersectional identities.

\begin{table}[b]
\centering
\begin{tabular}{llr}
\toprule
\textbf{Category} & \textbf{Attribute} & \textbf{Percentage (\%)} \\
\midrule
\multirow{2}{*}{Gender}
& Male & 62.99 \\
& Female & 37.01 \\
\midrule
\multirow{7}{*}{Race/Ethnicity}
& White & 56.43 \\
& Black & 3.73 \\
& Latino Hispanic & 4.41 \\
& East Asian & 11.86 \\
& Southeast Asian & 1.89 \\
& Indian & 4.20 \\
& Middle Eastern & 17.48 \\
\midrule
\multirow{9}{*}{Age}
& 0-2 & 0.05 \\
& 3-9 & 0.79 \\
& 10-19 & 4.67 \\
& 20-29 & 37.95 \\
& 30-39 & 25.83 \\
& 40-49 & 17.90 \\
& 50-59 & 8.45 \\
& 60-69 & 3.94 \\
& 70+ & 0.42 \\
\bottomrule
\end{tabular}
\vspace{-0.5em}
\caption{Demographic distribution analysis results on the HiFi-AVDF dataset.}
\label{tab:demographics}
\vspace{-0.2em}
\end{table}

\subsection{Other Datasets}
\noindent\textbf{LRS2}~\cite{lrs}.
LRS2 is a large-scale, unconstrained audio-visual dataset for speech recognition. It comprises 97k real videos sourced from British television, each paired with its corresponding audio track, enabling the modeling of both real human facial movements and their corresponding audio signals, and capturing the intrinsic audio-visual coherence.

\vspace{.5em}\noindent\textbf{FakeAVCeleb}~\cite{fakeavceleb}.
FakeAVCeleb is a deepfake detection dataset of 20,000 videos, comprising 500 real videos from VoxCeleb2~\cite{chung2018voxceleb2} and 19,500 deepfakes created via visual (FaceSwap~\cite{faceswap}, FSGAN~\cite{fsgan}, Wav2Lip~\cite{wav2lip}) and audio (SV2TTS~\cite{sv2tts}) manipulations. Based on which modalities are manipulated, the dataset can be categorized into four types: real video with fake audio (RVFA), fake video with real audio (FVRA), fake video with fake audio (FVFA), and real video with real audio (RVRA, i.e., the unaltered samples). Furthermore, the forgeries are generated using different combinations of manipulation techniques, as summarized in Tab.~\ref{tab:av_types}.
\begin{table}[b]
\centering
\renewcommand{\arraystretch}{1.05} 
\begin{adjustbox}{width=0.8\linewidth,center} 
\begin{tabular}{cc}
\toprule
Category & Generation Method \\
\midrule
RVFA        & SV2TTS \\
\hdashline
FVRA-FS     & FaceSwap \\
FVRA-GAN    & FaceSwapGAN \\
FVRA-WL     & Wav2Lip \\
\hdashline
FVFA-FS     & SV2TTS + FaceSwap \\
FVFA-GAN    & SV2TTS + FaceSwapGAN \\
FVFA-WL     & SV2TTS + Wav2Lip \\
\bottomrule
\end{tabular}
\end{adjustbox}
\vspace{-0.5em}
\caption{Overview of generation methods corresponding to different audio-visual manipulation categories in FakeAVCeleb.}
\label{tab:av_types}
\end{table}

\begin{table*}[t]
    \centering
    \begin{adjustbox}{width=0.95\linewidth,center}
    \begin{tabular}{r c *{10}{c} *{2}{c}}
    \toprule
    \multirow{2}{*}{Method} & \multicolumn{2}{c}{RVFA} & \multicolumn{2}{c}{FVRA-WL} & \multicolumn{2}{c}{FVFA-FS} & \multicolumn{2}{c}{FVFA-GAN} & \multicolumn{2}{c}{FVFA-WL} & \multicolumn{2}{c}{AVG} \\
    \cmidrule(lr){2-3} \cmidrule(lr){4-5} \cmidrule(lr){6-7} \cmidrule(lr){8-9} \cmidrule(lr){10-11} \cmidrule(lr){12-13}
     & AP & AUC & AP & AUC & AP & AUC & AP & AUC & AP & AUC & AP & AUC\\
    \midrule
    AV-DFD \cite{avdfd} & 74.9 & 73.3 & 97.0 & 97.4 & \underline{99.6} & \underline{99.7} & 58.4 & 55.4 & \textbf{100.} & \textbf{100.} & 88.8 & 88.1 \\
    AVAD (LRS2) \cite{avad} & 62.4 & 71.6 & 93.6 & 93.7 & 95.3 & 95.8 & 94.1 & \underline{94.3} & 93.8 & 94.1 & 94.2 & 94.5\\
    AVAD (LRS3) \cite{avad} & 70.7 & 80.5 & 91.1 & 93.0 & 91.0 & 92.3 & 91.6 & 92.7 & 91.4 & 93.1 & 91.3 & 92.8\\
    AVoiD-DF \cite{avoid-df} & 70.7 & 80.5 & 91.1 & 93.0 & 91.0 & 92.3 & 91.6 & 92.7 & 91.4 & 93.1 & 91.3 & 92.8\\
    AVFF \cite{oorloff2024avff} & 93.3 & 92.4 & 94.8 & 98.2 & \textbf{100.} & \textbf{100.} & \underline{99.9} & \textbf{100.} & \underline{99.4} & \underline{99.8} & 98.5 & 99.5\\
    AVPrompt \cite{avprompt} & \underline{97.1} & \underline{95.5} & \underline{99.9} & \underline{99.9} & \textbf{100.} & \textbf{100.} & \textbf{100.} & \textbf{100.} & \textbf{100.} & \textbf{100.} & \underline{99.4} & \underline{99.1}\\
    \midrule
    \textbf{HAVIC (Ours)} & \textbf{98.6} & \textbf{96.7} & \textbf{100.} & \textbf{100.} & \textbf{100.} & \textbf{100.} & \textbf{100.} & \textbf{100.} & \textbf{100.} & \textbf{100.} & \textbf{99.7} & \textbf{99.3}\\
    \bottomrule
    \end{tabular}
    \end{adjustbox}
    \caption{\textbf{Cross-manipulation generalization on FakeAVCeleb.} We evaluate the model on unseen manipulation types by training on four categories and testing on the held-out category. HAVIC consistently achieves superior performance across all manipulation types, demonstrating strong generalization to unseen deepfake generation methods.
    }
    \label{tab:cross_manip_eval}
    \vspace{-0.75em}
\end{table*}

\vspace{.5em}\noindent\textbf{KoDF}~\cite{kodf}.
KoDF is a large-scale talking-face deepfake dataset comprising 62,166 real videos and 175,776 fake videos generated using six synthesis algorithms: FaceSwap~\cite{faceswap}, DeepFaceLab~\cite{deepfacelab}, FaceSwapGAN~\cite{fsgan}, FOMM~\cite{fomm}, ATFHP~\cite{atfhp}, and Wav2Lip~\cite{wav2lip}. Following~\cite{oorloff2024avff,avad}, we use a subset of KoDF to evaluate the cross-dataset generalization (Tab. 3 in the main paper).

\section{Additional Results} \label{supp_sec:results}

\subsection{Cross-Manipulation Generalization.}
In addition to the intra-dataset evaluation reported on the FakeAVCeleb dataset (Tab. 2 in the main paper), we further conduct cross-manipulation experiments on the FakeAVCeleb dataset, following the protocols in \cite{oorloff2024avff,avad,avprompt}. The dataset is divided into five categories according to the specific deepfake generation algorithms: RVFA, FVRA-WL, FVFA-FS, FVFA-GAN, and FVFA-WL. In each experiment, we hold out one category for testing while training the model on the remaining four categories. This leave-one-type-out evaluation setup enables us to measure the model's ability to detect unseen manipulation methods. The results are summarized in Tab.~\ref{tab:cross_manip_eval}. HAVIC consistently outperforms all baseline methods across all manipulation types. This demonstrates that our model effectively captures generalizable audio-visual cues that transfer well to unseen deepfake generation methods. \looseness=-1

\subsection{Ablation on the absence of the audio modality.}
In practical scenarios, many videos may contain no audio or severely corrupted audio tracks. Most existing audio–visual detection methods heavily rely on both modalities, making them vulnerable when audio is missing. To evaluate the robustness of HAVIC under such conditions, we perform an ablation where the audio modality is entirely removed during testing. In this setting, we only use HAVIC's visual classification head for inference. For the two compared methods~\cite{oorloff2024avff,pia} that do not support audio-less input, we provide a silent audio track as a placeholder. As shown in Tab.~\ref{tab:ablation_no_audio}, all compared methods experience a notable drop in performance when audio is absent. In contrast, HAVIC maintains strong performance, showing that it can still perform reliable detection using only visual information.

\begin{table}[b]
    \centering
    \renewcommand{\arraystretch}{1.2}
    \begin{adjustbox}{width=0.95\linewidth,center}
        \begin{tabular}{l cc cc cc}
            \toprule
            \multirow{2}{*}{Method} 
            & \multicolumn{2}{c}{FakeAVCeleb} 
            & \multicolumn{2}{c}{KoDF} 
            & \multicolumn{2}{c}{HiFi-AVDF} \\
            \cmidrule(lr){2-3}
            \cmidrule(lr){4-5}
            \cmidrule(lr){6-7}
            & ACC & AUC & AP & AUC & AP & AUC \\
            \midrule
            AVFF~\cite{oorloff2024avff} & 84.9  & 86.5  & 77.2 & 78.4 & 56.2 & 55.3 \\
            PIA~\cite{pia} & 87.2  & 90.3 & 82.1 & 85.9 & 53.7 & 55.8 \\
            Visual Cls. of HAVIC   & 96.6 & 98.1 & 91.9 & 93.7 & 59.2 & 61.4 \\
            \midrule
            \textbf{HAVIC (Ours)} 
            & \textbf{99.8} & \textbf{99.9} 
            & \textbf{99.2} & \textbf{98.9} 
            & \textbf{75.4} & \textbf{75.7} \\
            \bottomrule
        \end{tabular}
    \end{adjustbox}
    \vspace{-0.5em}
    \caption{Ablation on the absence of the audio modality.}
    \label{tab:ablation_no_audio}
\end{table}

\subsection{Comparisons with other SSL methods.}
To further validate the effectiveness of our self-supervised learning (SSL) design in the Holistic Coherence Priors pre-training phase, we compare HAVIC with representative SSL methods from two perspectives.

\vspace{.4em}\noindent\textbf{MAE-based Pre-training.}
Since our method introduces hierarchical decoding and layer-wise supervision design beyond standard MAE, we compare HAVIC with several MAE variants to assess the benefit of these improvements.
Specifically, we substitute our design with two representative variants: VideoMAE~\cite{tong2022videomae} and HiCMAE~\cite{hicmae}, and pre-train the model under the same settings. As shown in Tab.~\ref{tab:ablation_MAE}, VideoMAE only uses the features from the last encoder layer for reconstruction, resulting in moderate performance. HiCMAE improves upon this by incorporating skip connections between the encoder and decoder, encouraging intermediate layers to learn more meaningful representations, which leads to better performance. Building on this, our method further introduces hierarchical decoding and layer-wise supervision, yielding additional performance gains.

\begin{figure*}[t]
    \centering
    \includegraphics[width=1.\linewidth]{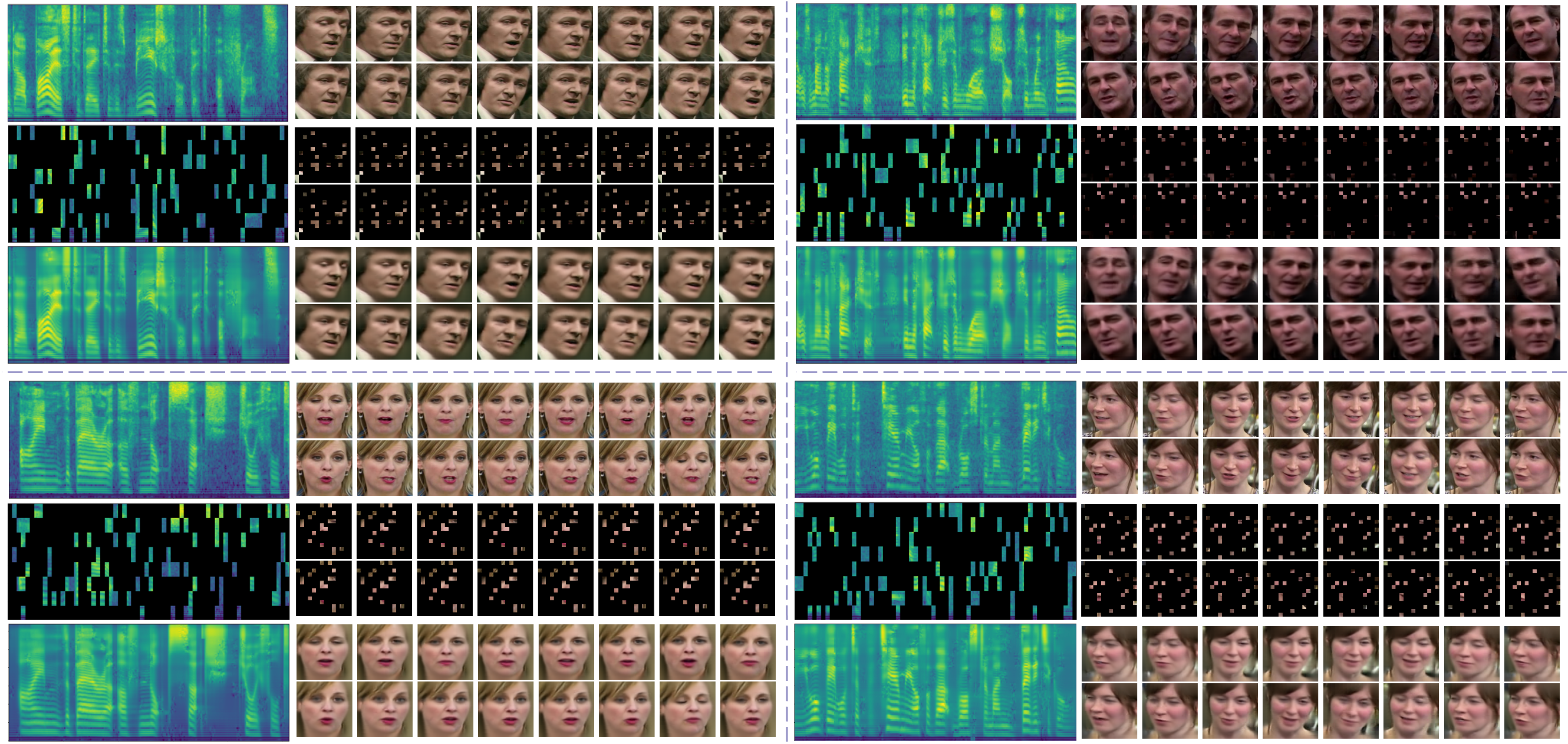 }
    \vspace{-1.5em}
    \caption{\textbf{Modality-Specific Hierarchical Reconstruction visualizations}. For each clip, the first row shows the original audio spectrograms and visual frames, while the second and third rows depict the masked inputs and the corresponding reconstructions from the final decoder layer, respectively. Details of the reconstructions can be seen by zooming in.}
    \label{fig:mask_viz}
    \vspace{-1em}
\end{figure*}

\vspace{.4em}\noindent\textbf{Audio-Visual Contrastive Learning.}
To evaluate the contribution of our fine-grained audio-visual contrastive learning, we conduct two ablation studies targeting its key components: temporal segmentation and the soft negative mechanism. Previous contrastive methods~\cite{oorloff2024avff,cav-mae} treat an entire audio–video pair as a single unit, ignoring temporal structure. We first replace our segment-level formulation with a video-level contrastive loss by globally pooling both modalities to assess the impact of temporal segmentation. Second, we remove the soft negative mechanism. As shown in Tab.~\ref{tab:ablation_AVCL}, both components contribute to the model's performance, demonstrating the effectiveness of our designs.

\begin{table}[b]
    \centering
    \renewcommand{\arraystretch}{1.2}
    \begin{adjustbox}{width=0.95\linewidth,center}
        \begin{tabular}{l cc cc cc}
            \toprule
            \multirow{2}{*}{Method} 
            & \multicolumn{2}{c}{FakeAVCeleb} 
            & \multicolumn{2}{c}{KoDF} 
            & \multicolumn{2}{c}{HiFi-AVDF} \\
            \cmidrule(lr){2-3}
            \cmidrule(lr){4-5}
            \cmidrule(lr){6-7}
            & ACC & AUC & AP & AUC & AP & AUC \\
            \midrule
            VideoMAE~\cite{tong2022videomae} & 98.4  & 99.1  & 94.5 & 95.2 & 69.1 & 71.3 \\
            HiCMAE~\cite{hicmae}  & 99.3 & 99.5 & 96.7 & 96.3 & 71.8 & 73.5 \\
            \midrule
            \textbf{HAVIC (Ours)} 
            & \textbf{99.8} & \textbf{99.9} 
            & \textbf{99.2} & \textbf{98.9} 
            & \textbf{75.4} & \textbf{75.7} \\
            \bottomrule
        \end{tabular}
    \end{adjustbox}
    \vspace{-0.5em}
    \caption{Comparison with MAE-based Pre-training Variants.}
    \label{tab:ablation_MAE}
\end{table}

\begin{table}[b]
    \centering
    \renewcommand{\arraystretch}{1.2}
    \begin{adjustbox}{width=0.95\linewidth,center}
        \begin{tabular}{l cc cc cc}
            \toprule
            \multirow{2}{*}{Method} 
            & \multicolumn{2}{c}{FakeAVCeleb} 
            & \multicolumn{2}{c}{KoDF} 
            & \multicolumn{2}{c}{HiFi-AVDF} \\
            \cmidrule(lr){2-3}
            \cmidrule(lr){4-5}
            \cmidrule(lr){6-7}
            & ACC & AUC & AP & AUC & AP & AUC \\
            \midrule
            w/o temporal segments    & 98.7 & 98.9 & 96.5 & 96.8 & 69.6 & 70.8 \\
            w/o soft negative mechanism   & 99.4 & 99.5 & 98.4 & 98.1 & 74.0 & 75.1 \\
            \midrule
            \textbf{HAVIC (Ours)} 
            & \textbf{99.8} & \textbf{99.9} 
            & \textbf{99.2} & \textbf{98.9} 
            & \textbf{75.4} & \textbf{75.7} \\
            \bottomrule
        \end{tabular}
    \end{adjustbox}
    \vspace{-1em}
    \caption{Ablation study on key components of Fine-grained Audio-Visual Contrastive Learning.}
    \label{tab:ablation_AVCL}
\end{table}

\subsection{Hyperparameter Sensitivity of Loss Weight.}
To assess the influence of the loss weight in the Fine-grained Audio-Visual Contrastive Loss, we experiment with different weighting values. As shown in Tab.~\ref{tab:ablation_loss_weight}, a weight around $\lambda_{cl}=0.01$ yields the best performance across all datasets.
A smaller weight weakens the contrastive learning signal, making the model insufficiently align audio–visual features, whereas an excessively large weight overwhelms other objectives and disrupts the overall optimization balance.

\begin{table}[b]
\centering
\renewcommand{\arraystretch}{1.1}
\resizebox{0.95\linewidth}{!}{%
\begin{tabular}{c cc cc cc}
\toprule
\multirow{2}{*}{Loss weight $\lambda_{cl}$} 
& \multicolumn{2}{c}{FakeAVCeleb} 
& \multicolumn{2}{c}{KoDF} 
& \multicolumn{2}{c}{HiFi-AVDF} \\
\cmidrule(lr){2-3} \cmidrule(lr){4-5} \cmidrule(lr){6-7}
& ACC & AUC & AP & AUC & AP & AUC \\
\midrule
0.001  & 99.8 & 99.8 & 97.3 & 96.7 & 72.3 & 67.1 \\
0.005  & 99.8 & 99.9 & 98.5 & 99.0 & 74.7 & 74.4 \\
\rowcolor{orange} 0.01   & 99.8 & 99.9 & 99.2 & 98.8 & 75.4 & 75.7 \\
0.05   & 99.8 & 99.8 & 98.3 & 98.4 & 73.9 & 75.5 \\
0.1    & 99.6 & 99.7 & 97.8 & 97.9 & 72.4 & 73.8 \\
0.5    & 99.3 & 99.8 & 96.5 & 96.7 & 72.8 & 71.6 \\
1      & 98.7 & 99.1 & 96.2 & 96.9 & 70.2 & 69.8 \\
\bottomrule
\end{tabular}%
}
\caption{Ablation study on loss weight for Fine-grained Audio-visual Contrastive Loss.}
\label{tab:ablation_loss_weight}
\end{table}

\begin{figure*}[t]
    \centering
    \includegraphics[width=1.\linewidth]{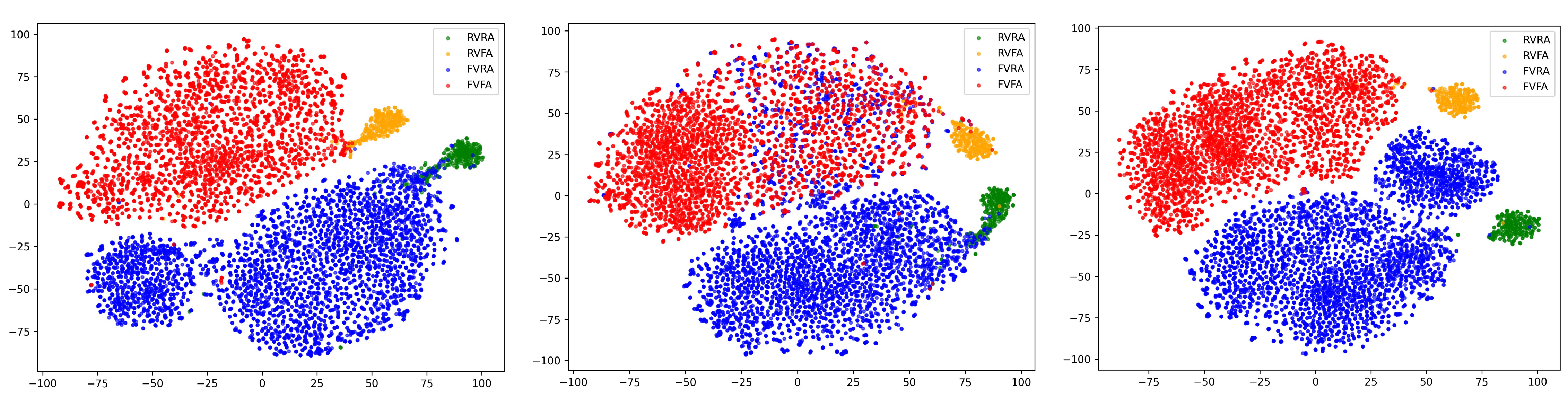}
    \vspace{-1.5em}
    \caption{\textbf{t-SNE visualization of the learned audio-visual embeddings}. From left to right: (1) embeddings without Adaptive Aggregation, (2) embeddings without Auxiliary Classifiers, and (3) embeddings of the complete model.}
    \label{fig:tnse}
    \vspace{-1em}
\end{figure*}

\subsection{Analysis of Model Complexity.}
We analyze the trade-off between model performance and computational efficiency. Tab.~\ref{tab:model_complexity} compares the number of parameters, throughput, and performance on HiFi-AVDF dataset of representative MAE-based pre-training models. Our proposed HAVIC achieves the highest AP and AUC while maintaining competitive throughput and a moderate number of parameters, demonstrating an efficient balance between accuracy and computational cost.

Furthermore, we conduct an ablation study on the components of Hierarchical Adaptive Aggregation Classification phase to examine their impact on model complexity and efficiency. Tab.~\ref{tab:HAA_ablation_params} shows that both adaptive aggregation and auxiliary classifiers contribute to improved performance, with a slight reduction in throughput as more components are added. This analysis highlights the trade-off between incorporating advanced modeling components and maintaining efficient inference speed.

\begin{table}[t]
\centering
\renewcommand{\arraystretch}{1.2}
\resizebox{0.95\linewidth}{!}{
\begin{tabular}{lcccc}
\toprule
Method & Parameters (M) & Throughput (samples/s) & AP & AUC \\
\midrule
VideoMAE~\cite{tong2022videomae} & 235.1 & 74.7 & 69.1 & 71.3 \\
HiCMAE~\cite{hicmae} & 241.0 & 72.5 & 71.8 & 73.5 \\
AVFF~\cite{oorloff2024avff} & 196.8 & 34.2 & 66.0 & 65.2 \\
\midrule
\textbf{HAVIC (Ours)} & 243.1 & 66.8 & 75.4 & 75.7 \\
\bottomrule
\end{tabular}
}
\caption{Comparison of model complexity and performance. The number of parameters and inference throughput are reported for the pre-training models, while AP and AUC are evaluated on the HiFi-AVDF dataset to assess detection performance.}
\label{tab:model_complexity}
\vspace{-1em}
\end{table}

\begin{table}[h]
\centering
\renewcommand{\arraystretch}{1.2}
\begin{adjustbox}{width=0.95\linewidth,center}
\begin{tabular}{cccccc}
\toprule
\makecell{Adaptive\\ Aggregation} & \makecell{Auxiliary\\ Classifiers} & Parameters (M) & Throughput (samples/s) & AP & AUC \\
\midrule
\xmark & \xmark & 223.2 & 67.0 & 65.3 & 67.6 \\
\cmark  & \xmark & 224.2 & 65.4 & 67.7 & 71.8 \\
\xmark & \cmark & 227.9 & 64.2 & 71.2 & 72.9 \\
\cmark  & \cmark & 228.9 & 60.8 & 75.4 & 75.7 \\
\bottomrule
\end{tabular}
\end{adjustbox}
\vspace{-0.5em}
\caption{Impact of components in the Hierarchical Adaptive Aggregation Classification phase on model size, throughput, and performance on the HiFi-AVDF dataset. \xmark and \cmark indicate the exclusion and inclusion of each component, respectively.}
\label{tab:HAA_ablation_params}
\vspace{-1em}
\end{table}

\subsection{Multiple Runs}
In the main paper, we report performance metrics averaged over multiple runs with different random seeds to mitigate randomness and enable reliable comparisons across three benchmark datasets. 
To provide a detailed view of variability, Tab.~\ref{tab:ablation_Multiple_Runs} presents the results of five individual runs for each dataset, along with their mean and standard deviation.

\begin{table}[b]
    \centering
    \renewcommand{\arraystretch}{1.2}
    \begin{adjustbox}{width=0.95\linewidth,center}
        \begin{tabular}{c cc cc cc}
            \toprule
            \multirow{2}{*}{Multiple Runs} 
            & \multicolumn{2}{c}{FakeAVCeleb} 
            & \multicolumn{2}{c}{KoDF} 
            & \multicolumn{2}{c}{HiFi-AVDF} \\
            \cmidrule(lr){2-3}
            \cmidrule(lr){4-5}
            \cmidrule(lr){6-7}
            & ACC & AUC & AP & AUC & AP & AUC \\
            \midrule
             (i) & 99.81 & 99.95 & 99.32 & 98.86 &75.93  &  76.04  \\
             (ii) & 99.84 & 99.96 & 99.39 & 99.24 & 75.25 &  75.85  \\
             (iii) & 99.90 & 99.99 & 98.96 & 98.63 & 75.06 & 74.91  \\
             (iv)  & 99.86 & 99.98 & 98.79 & 98.87 & 74.19 & 75.58 \\
             (v)  & 99.79 & 99.93 & 99.59 & 99.10 & 76.62 & 76.49 \\
            \midrule
            Mean & 99.84 & 99.96 & 99.21 & 98.94 & 75.41 & 75.77 \\
            std & 0.04 & 0.02 & 0.33 & 0.24 & 0.92 & 0.59\\
             
            \bottomrule
        \end{tabular}
    \end{adjustbox}
    \vspace{-0.5em}
    \caption{Performance across 5 runs on three benchmark datasets.}
    \label{tab:ablation_Multiple_Runs}
\end{table}

\subsection{Qualitative Analysis}

\vspace{.4em}\noindent\textbf{Modality-Specific Reconstruction Visualizations.} 
We present visualizations of the Modality-Specific Hierarchical Reconstruction in Fig.~\ref{fig:mask_viz}. 
Video clips are randomly selected from the unseen test set. For each clip, the first row shows the original audio spectrogram and 16 visual frames, while the second and third rows depict the masked inputs and the corresponding reconstructions, respectively. All reconstruction results are produced by the final layer of the decoders.

The reconstructions closely resemble the original inputs, successfully recovering the overall structure of both modalities. 
Although the reconstructions are slightly smoother than the ground truth as a result of the high masking ratio, HAVIC still restores informative and coherent patterns.
These results indicate that HAVIC can effectively infer missing content from limited visible information while capturing meaningful features, demonstrating that it has learned robust intra-modal structural coherence priors that benefit downstream audio-visual deepfake detection.

\vspace{.4em}\noindent\textbf{Visualization of Embedding Space with t-SNE.}
To further analyze the impact of the Hierarchical Adaptive Aggregation Classification phase (Tab. 6 in the main paper), we visualize the learned audio-visual embeddings using t-SNE~\cite{tSNE}. As shown in the Fig.~\ref{fig:tnse}, in the leftmost plot (w/o Adaptive Aggregation), the clusters exhibit some overlap, likely because average feature aggregation weakens the discriminative information of certain strong features. In the middle plot (w/o Auxiliary Classifiers), the overlap between forged samples is more pronounced, particularly for FVFA and FVRA, as the model is only trained to output overall real/fake predictions, limiting its ability to capture modality-specific discrepancies. In contrast, the rightmost plot shows embeddings from our full model, where the clusters are clearly separated, highlighting the effectiveness of our proposed Hierarchical Adaptive Aggregation Classification phase in producing discriminative features for deepfake detection.

\subsection{Failure Case Analysis}
\begin{figure}[h]
    \centering
    \includegraphics[width=0.95\linewidth]{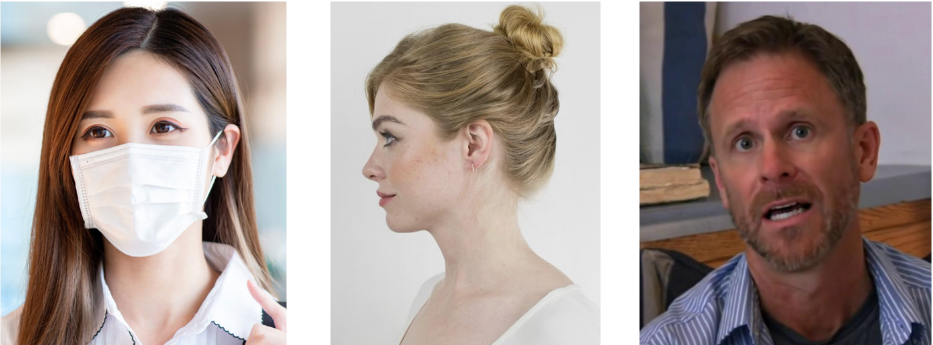}
     \vspace{-0.5em}
    \caption{Examples of challenging failure cases. From left to right:  (1) masked faces, (2) profile or side faces, and (3) highly realistic forgeries.}
    \label{fig:Failure-Case}
    \vspace{-1em}
\end{figure}

Despite the overall strong performance of our proposed HAVIC, there exist certain scenarios in which the model fails to correctly detect forged content. Through qualitative examination, we identify three main types of challenging cases: (1) \textbf{masked faces}, where a large portion of the face is occluded, reducing the quality of visual cues and hindering reliable audio-visual correspondence; (2) \textbf{profile or side faces}, which reduce the visibility of distinctive facial features and limit the usable visual information; and (3) \textbf{highly realistic forgeries}, where the generated audio-visual cues are nearly indistinguishable from authentic data. Fig.~\ref{fig:Failure-Case} illustrates representative examples from each category.

{
    \small
    \bibliographystyle{ieeenat_fullname}
    \bibliography{main}

@String(CVPR= {IEEE Conf. Comput. Vis. Pattern Recog.})

@String(TOG= {ACM Trans. Graph.})

@String(ICASSP=	{ICASSP})

@String(AAAI = {AAAI})

@String(CVPR  = {CVPR})

@String(TOG   = {ACM TOG})

@inproceedings{MAE,
  title={Masked autoencoders are scalable vision learners},
  author={He, Kaiming and Chen, Xinlei and Xie, Saining and Li, Yanghao and Doll{\'a}r, Piotr and Girshick, Ross},
  booktitle={Proceedings of the IEEE/CVF conference on computer vision and pattern recognition},
  pages={16000--16009},
  year={2022}
}

@article{tong2022videomae,
  title={Videomae: Masked autoencoders are data-efficient learners for self-supervised video pre-training},
  author={Tong, Zhan and Song, Yibing and Wang, Jue and Wang, Limin},
  journal={Advances in neural information processing systems},
  volume={35},
  pages={10078--10093},
  year={2022}
}

@article{audiomae,
  title={Masked autoencoders that listen},
  author={Huang, Po-Yao and Xu, Hu and Li, Juncheng and Baevski, Alexei and Auli, Michael and Galuba, Wojciech and Metze, Florian and Feichtenhofer, Christoph},
  journal={Advances in Neural Information Processing Systems},
  volume={35},
  pages={28708--28720},
  year={2022}
}

@inproceedings{oorloff2024avff,
  title={Avff: Audio-visual feature fusion for video deepfake detection},
  author={Oorloff, Trevine and Koppisetti, Surya and Bonettini, Nicol{\`o} and Solanki, Divyaraj and Colman, Ben and Yacoob, Yaser and Shahriyari, Ali and Bharaj, Gaurav},
  booktitle={Proceedings of the IEEE/CVF Conference on Computer Vision and Pattern Recognition},
  pages={27102--27112},
  year={2024}
}

@article{AVgraph,
  title={Fine-grained multimodal deepfake classification via heterogeneous graphs},
  author={Yin, Qilin and Lu, Wei and Cao, Xiaochun and Luo, Xiangyang and Zhou, Yicong and Huang, Jiwu},
  journal={International Journal of Computer Vision},
  volume={132},
  number={11},
  pages={5255--5269},
  year={2024},
  publisher={Springer}
}

@article{cav-mae,
  title={Contrastive audio-visual masked autoencoder},
  author={Gong, Yuan and Rouditchenko, Andrew and Liu, Alexander H and Harwath, David and Karlinsky, Leonid and Kuehne, Hilde and Glass, James},
  journal={arXiv preprint arXiv:2210.07839},
  year={2022}
}

@article{hicmae,
  title={Hicmae: Hierarchical contrastive masked autoencoder for self-supervised audio-visual emotion recognition},
  author={Sun, Licai and Lian, Zheng and Liu, Bin and Tao, Jianhua},
  journal={Information Fusion},
  volume={108},
  pages={102382},
  year={2024},
  publisher={Elsevier}
}

@inproceedings{avprompt,
  title={Multi-modal Deepfake Detection via Multi-task Audio-Visual Prompt Learning},
  author={Miao, Hui and Guo, Yuanfang and Liu, Zeming and Wang, Yunhong},
  booktitle={Proceedings of the AAAI Conference on Artificial Intelligence},
  volume={39},
  number={1},
  pages={612--621},
  year={2025}
}

@article{fakeavceleb,
  title={FakeAVCeleb: A novel audio-video multimodal deepfake dataset},
  author={Khalid, Hasam and Tariq, Shahroz and Kim, Minha and Woo, Simon S},
  journal={arXiv preprint arXiv:2108.05080},
  year={2021}
}

@inproceedings{kodf,
  title={Kodf: A large-scale korean deepfake detection dataset},
  author={Kwon, Patrick and You, Jaeseong and Nam, Gyuhyeon and Park, Sungwoo and Chae, Gyeongsu},
  booktitle={Proceedings of the IEEE/CVF international conference on computer vision},
  pages={10744--10753},
  year={2021}
}

@inproceedings{avad,
  title={Self-Supervised Video Forensics by Audio-Visual Anomaly Detection},
  author={Feng, Chao and Chen, Ziyang and Owens, Andrew},
  booktitle={Proceedings of the IEEE/CVF Conference on Computer Vision and Pattern Recognition},
  pages={10491--10503},
  year={2023}
}

@article{lrs,
  title={Deep audio-visual speech recognition},
  author={Afouras, Triantafyllos and Chung, Joon Son and Senior, Andrew and Vinyals, Oriol and Zisserman, Andrew},
  journal={IEEE transactions on pattern analysis and machine intelligence},
  volume={44},
  number={12},
  pages={8717--8727},
  year={2018},
  publisher={IEEE}
}

@inproceedings{facexzoo,
  title={Facex-zoo: A pytorch toolbox for face recognition},
  author={Wang, Jun and Liu, Yinglu and Hu, Yibo and Shi, Hailin and Mei, Tao},
  booktitle={Proceedings of the 29th ACM international conference on Multimedia},
  pages={3779--3782},
  year={2021}
}

@inproceedings{marlin,
  title={Marlin: Masked autoencoder for facial video representation learning},
  author={Cai, Zhixi and Ghosh, Shreya and Stefanov, Kalin and Dhall, Abhinav and Cai, Jianfei and Rezatofighi, Hamid and Haffari, Reza and Hayat, Munawar},
  booktitle={Proceedings of the IEEE/CVF conference on computer vision and pattern recognition},
  pages={1493--1504},
  year={2023}
}

@inproceedings{mrdf,
  title={Cross-modality and within-modality regularization for audio-visual deepfake detection},
  author={Zou, Heqing and Shen, Meng and Hu, Yuchen and Chen, Chen and Chng, Eng Siong and Rajan, Deepu},
  booktitle={ICASSP 2024-2024 IEEE International Conference on Acoustics, Speech and Signal Processing (ICASSP)},
  pages={4900--4904},
  year={2024},
  organization={IEEE}
}

@inproceedings{Xception,
  title={Faceforensics++: Learning to detect manipulated facial images},
  author={Rossler, Andreas and Cozzolino, Davide and Verdoliva, Luisa and Riess, Christian and Thies, Justus and Nie{\ss}ner, Matthias},
  booktitle={Proceedings of the IEEE/CVF international conference on computer vision},
  pages={1--11},
  year={2019}
}

@inproceedings{realforensics,
  title={Leveraging Real Talking Faces via Self-Supervision for Robust Forgery Detection},
  author={Haliassos, Alexandros and Mira, Rodrigo and Petridis, Stavros and Pantic, Maja},
  booktitle={Proceedings of the IEEE/CVF Conference on Computer Vision and Pattern Recognition},
  pages={14950--14962},
  year={2022}
}

@inproceedings{lip-forensics,
  title={Lips don't lie: A generalisable and robust approach to face forgery detection},
  author={Haliassos, Alexandros and Vougioukas, Konstantinos and Petridis, Stavros and Pantic, Maja},
  booktitle={Proceedings of the IEEE/CVF conference on computer vision and pattern recognition},
  pages={5039--5049},
  year={2021}
}

@inproceedings{ftcn,
  title={Exploring temporal coherence for more general video face forgery detection},
  author={Zheng, Yinglin and Bao, Jianmin and Chen, Dong and Zeng, Ming and Wen, Fang},
  booktitle={Proceedings of the IEEE/CVF international conference on computer vision},
  pages={15044--15054},
  year={2021}
}

@article{tSNE,
  title={Visualizing data using t-SNE},
  author={Maaten, Laurens van der and Hinton, Geoffrey},
  journal={Journal of machine learning research},
  volume={9},
  number={Nov},
  pages={2579--2605},
  year={2008}
}

@article{dfdc,
  title={The deepfake detection challenge (dfdc) dataset},
  author={Dolhansky, Brian and Bitton, Joanna and Pflaum, Ben and Lu, Jikuo and Howes, Russ and Wang, Menglin and Ferrer, Cristian Canton},
  journal={arXiv preprint arXiv:2006.07397},
  year={2020}
}

@article{faceforensics,
  title={Faceforensics: A large-scale video dataset for forgery detection in human faces},
  author={R{\"o}ssler, Andreas and Cozzolino, Davide and Verdoliva, Luisa and Riess, Christian and Thies, Justus and Nie{\ss}ner, Matthias},
  journal={arXiv preprint arXiv:1803.09179},
  year={2018}
}

@inproceedings{avdfd,
  title={Joint audio-visual deepfake detection},
  author={Zhou, Yipin and Lim, Ser-Nam},
  booktitle={Proceedings of the IEEE/CVF international conference on computer vision},
  pages={14800--14809},
  year={2021}
}

@article{avoid-df,
  title={Avoid-df: Audio-visual joint learning for detecting deepfake},
  author={Yang, Wenyuan and Zhou, Xiaoyu and Chen, Zhikai and Guo, Bofei and Ba, Zhongjie and Xia, Zhihua and Cao, Xiaochun and Ren, Kui},
  journal={IEEE Transactions on Information Forensics and Security},
  volume={18},
  pages={2015--2029},
  year={2023},
  publisher={IEEE}
}

@article{speechforensics,
  title={SpeechForensics: Audio-visual speech representation learning for face forgery detection},
  author={Liang, Yachao and Yu, Min and Li, Gang and Jiang, Jianguo and Li, Boquan and Yu, Feng and Zhang, Ning and Meng, Xiang and Huang, Weiqing},
  journal={Advances in Neural Information Processing Systems},
  volume={37},
  pages={86124--86144},
  year={2024}
}

@inproceedings{avh-align,
  title={Circumventing shortcuts in audio-visual deepfake detection datasets with unsupervised learning},
  author={Smeu, Stefan and Boldisor, Dragos-Alexandru and Oneata, Dan and Oneata, Elisabeta},
  booktitle={Proceedings of the Computer Vision and Pattern Recognition Conference},
  pages={18815--18825},
  year={2025}
}

@article{lipfd,
  title={Lips are lying: Spotting the temporal inconsistency between audio and visual in lip-syncing deepfakes},
  author={Liu, Weifeng and She, Tianyi and Liu, Jiawei and Li, Boheng and Yao, Dongyu and Wang, Run},
  journal={Advances in Neural Information Processing Systems},
  volume={37},
  pages={91131--91155},
  year={2024}
}

@inproceedings{pia,
  title={PIA: Deepfake Detection Using Phoneme-Temporal and Identity-Dynamic Analysis},
  author={Datta, Soumyya Kanti and Ranga, Tanvi and Sun, Chengzhe and Lyu, Siwei},
  booktitle={Proceedings of the IEEE/CVF International Conference on Computer Vision},
  pages={1596--1606},
  year={2025}
}

@inproceedings{phoneme_viseme_mismatch,
  title={Detecting deep-fake videos from phoneme-viseme mismatches},
  author={Agarwal, Shruti and Farid, Hany and Fried, Ohad and Agrawala, Maneesh},
  booktitle={Proceedings of the IEEE/CVF conference on computer vision and pattern recognition workshops},
  pages={660--661},
  year={2020}
}

@article{cViTforDFD,
  title={Deepfake video detection using convolutional vision transformer},
  author={Wodajo, Deressa and Atnafu, Solomon},
  journal={arXiv preprint arXiv:2102.11126},
  year={2021}
}

@inproceedings{face_x_ray,
  title={Face x-ray for more general face forgery detection},
  author={Li, Lingzhi and Bao, Jianmin and Zhang, Ting and Yang, Hao and Chen, Dong and Wen, Fang and Guo, Baining},
  booktitle={Proceedings of the IEEE/CVF conference on computer vision and pattern recognition},
  pages={5001--5010},
  year={2020}
}

@inproceedings{dang2020detection,
  title={On the detection of digital face manipulation},
  author={Dang, Hao and Liu, Feng and Stehouwer, Joel and Liu, Xiaoming and Jain, Anil K},
  booktitle={Proceedings of the IEEE/CVF Conference on Computer Vision and Pattern recognition},
  pages={5781--5790},
  year={2020}
}

@inproceedings{Multi_attentional_deepfake_detection,
  title={Multi-attentional deepfake detection},
  author={Zhao, Hanqing and Zhou, Wenbo and Chen, Dongdong and Wei, Tianyi and Zhang, Weiming and Yu, Nenghai},
  booktitle={Proceedings of the IEEE/CVF conference on computer vision and pattern recognition},
  pages={2185--2194},
  year={2021}
}

@inproceedings{high-frequency,
  title={Generalizing face forgery detection with high-frequency features},
  author={Luo, Yuchen and Zhang, Yong and Yan, Junchi and Liu, Wei},
  booktitle={Proceedings of the IEEE/CVF conference on computer vision and pattern recognition},
  pages={16317--16326},
  year={2021}
}

@inproceedings{wilddeepfake,
  title={Wilddeepfake: A challenging real-world dataset for deepfake detection},
  author={Zi, Bojia and Chang, Minghao and Chen, Jingjing and Ma, Xingjun and Jiang, Yu-Gang},
  booktitle={Proceedings of the 28th ACM international conference on multimedia},
  pages={2382--2390},
  year={2020}
}

@inproceedings{avdeepfake1m,
  title={AV-Deepfake1M: A large-scale LLM-driven audio-visual deepfake dataset},
  author={Cai, Zhixi and Ghosh, Shreya and Adatia, Aman Pankaj and Hayat, Munawar and Dhall, Abhinav and Gedeon, Tom and Stefanov, Kalin},
  booktitle={Proceedings of the 32nd ACM International Conference on Multimedia},
  pages={7414--7423},
  year={2024}
}

@inproceedings{df-platter,
  title={Df-platter: Multi-face heterogeneous deepfake dataset},
  author={Narayan, Kartik and Agarwal, Harsh and Thakral, Kartik and Mittal, Surbhi and Vatsa, Mayank and Singh, Richa},
  booktitle={Proceedings of the IEEE/CVF Conference on Computer Vision and Pattern Recognition},
  pages={9739--9748},
  year={2023}
}

@inproceedings{faceswap,
  title={Fast face-swap using convolutional neural networks},
  author={Korshunova, Iryna and Shi, Wenzhe and Dambre, Joni and Theis, Lucas},
  booktitle={Proceedings of the IEEE international conference on computer vision},
  pages={3677--3685},
  year={2017}
}

@inproceedings{fsgan,
  title={Fsgan: Subject agnostic face swapping and reenactment},
  author={Nirkin, Yuval and Keller, Yosi and Hassner, Tal},
  booktitle={Proceedings of the IEEE/CVF international conference on computer vision},
  pages={7184--7193},
  year={2019}
}

@inproceedings{wav2lip,
  title={A lip sync expert is all you need for speech to lip generation in the wild},
  author={Prajwal, KR and Mukhopadhyay, Rudrabha and Namboodiri, Vinay P and Jawahar, CV},
  booktitle={Proceedings of the 28th ACM international conference on multimedia},
  pages={484--492},
  year={2020}
}

@article{sv2tts,
  title={Transfer learning from speaker verification to multispeaker text-to-speech synthesis},
  author={Jia, Ye and Zhang, Yu and Weiss, Ron and Wang, Quan and Shen, Jonathan and Ren, Fei and Nguyen, Patrick and Pang, Ruoming and Lopez Moreno, Ignacio and Wu, Yonghui and others},
  journal={Advances in neural information processing systems},
  volume={31},
  year={2018}
}

@inproceedings{talklip,
  title={Seeing what you said: Talking face generation guided by a lip reading expert},
  author={Wang, Jiadong and Qian, Xinyuan and Zhang, Malu and Tan, Robby T and Li, Haizhou},
  booktitle={Proceedings of the IEEE/CVF Conference on Computer Vision and Pattern Recognition},
  pages={14653--14662},
  year={2023}
}

@inproceedings{yourtts,
  title={Yourtts: Towards zero-shot multi-speaker tts and zero-shot voice conversion for everyone},
  author={Casanova, Edresson and Weber, Julian and Shulby, Christopher D and Junior, Arnaldo Candido and G{\"o}lge, Eren and Ponti, Moacir A},
  booktitle={International conference on machine learning},
  pages={2709--2720},
  year={2022},
  organization={PMLR}
}

@inproceedings{vits,
  title={Conditional variational autoencoder with adversarial learning for end-to-end text-to-speech},
  author={Kim, Jaehyeon and Kong, Jungil and Son, Juhee},
  booktitle={International Conference on Machine Learning},
  pages={5530--5540},
  year={2021},
  organization={PMLR}
}

@article{deepfacelab,
  title={DeepFaceLab: Integrated, flexible and extensible face-swapping framework},
  author={Perov, Ivan and Gao, Daiheng and Chervoniy, Nikolay and Liu, Kunlin and Marangonda, Sugasa and Um{\'e}, Chris and Facenheim, Carl Shift and RP, Luis and Jiang, Jian and Zhang, Sheng and others},
  journal={arXiv preprint arXiv:2005.05535},
  year={2020}
}

@inproceedings{evp,
  title={Audio-driven emotional video portraits},
  author={Ji, Xinya and Zhou, Hang and Wang, Kaisiyuan and Wu, Wayne and Loy, Chen Change and Cao, Xun and Xu, Feng},
  booktitle={Proceedings of the IEEE/CVF conference on computer vision and pattern recognition},
  pages={14080--14089},
  year={2021}
}

@inproceedings{pc-avs,
  title={Pose-controllable talking face generation by implicitly modularized audio-visual representation},
  author={Zhou, Hang and Sun, Yasheng and Wu, Wayne and Loy, Chen Change and Wang, Xiaogang and Liu, Ziwei},
  booktitle={Proceedings of the IEEE/CVF conference on computer vision and pattern recognition},
  pages={4176--4186},
  year={2021}
}

@article{voice-replay,
  title={The ASVspoof 2017 challenge: Assessing the limits of replay spoofing attack detection},
  author={Kinnunen, Tomi and Sahidullah, Md and Delgado, H{\'e}ctor and Todisco, Massimiliano and Evans, Nicholas and Yamagishi, Junichi and Lee, Kong Aik},
  year={2017},
  publisher={ISCA (the International Speech Communication Association)}
}

@article{av-exemplar-autoencoders,
  title={Unsupervised audiovisual synthesis via exemplar autoencoders},
  author={Deng, Kangle and Bansal, Aayush and Ramanan, Deva},
  journal={arXiv preprint arXiv:2001.04463},
  year={2020}
}

@article{fomm,
  title={First order motion model for image animation},
  author={Siarohin, Aliaksandr and Lathuili{\`e}re, St{\'e}phane and Tulyakov, Sergey and Ricci, Elisa and Sebe, Nicu},
  journal={Advances in neural information processing systems},
  volume={32},
  year={2019}
}

@article{atfhp,
  title={Audio-driven talking face video generation with learning-based personalized head pose},
  author={Yi, Ran and Ye, Zipeng and Zhang, Juyong and Bao, Hujun and Liu, Yong-Jin},
  journal={arXiv preprint arXiv:2002.10137},
  year={2020}
}

@inproceedings{face2face,
  title={Face2face: Real-time face capture and reenactment of rgb videos},
  author={Thies, Justus and Zollhofer, Michael and Stamminger, Marc and Theobalt, Christian and Nie{\ss}ner, Matthias},
  booktitle={Proceedings of the IEEE conference on computer vision and pattern recognition},
  pages={2387--2395},
  year={2016}
}

@article{neuraltextures,
  title={Deferred neural rendering: Image synthesis using neural textures},
  author={Thies, Justus and Zollh{\"o}fer, Michael and Nie{\ss}ner, Matthias},
  journal={Acm Transactions on Graphics (TOG)},
  volume={38},
  number={4},
  pages={1--12},
  year={2019},
  publisher={ACM New York, NY, USA}
}

@misc{deepfakes,
  title = {Deepfakes},
  howpublished ={\url{https://github.com/deepfakes/faceswap}},
  note = {Accessed: 2025-11-03}
}

@misc{DALL-E,
  title = {DALL-E},
  howpublished ={\url{https://openai.com/index/dall-e-2/}},
  note = {Accessed: 2025-11-03}
}

@misc{sora2,
  author = {OpenAI},
  title = {Sora 2},
  howpublished = {\url{https://openai.com/index/sora-2/}},
  note = {Accessed: 2025-11-03}
}

@misc{veo3.1,
  author = {Google},
  title = {Veo 3.1},
  howpublished = {\url{https://aistudio.google.com/models/veo-3}},
  note = {Accessed: 2025-11-03}
}

@misc{pixverse,
  author = {AISphere},
  title = {PixVerse},
  howpublished = {\url{https://app.pixverse.ai/home}},
  note = {Accessed: 2025-11-03}
}

@misc{seedance,
  author = {Bytedance},
  title = {Seedance},
  howpublished = {\url{https://seed.bytedance.com/en/seedance}},
  note = {Accessed: 2025-11-03}
}

@misc{wan2.5,
  author = {Alibaba},
  title = {WAN},
  howpublished = {\url{https://tongyi.aliyun.com/wan/}},
  note = {Accessed: 2025-11-03}
}

@misc{kling2.5,
  author = {kuaishou},
  title = {Kling},
  howpublished = {\url{https://klingai.com/}},
  note = {Accessed: 2025-11-03}
}

@article{faceshifter,
  title={Faceshifter: Towards high fidelity and occlusion aware face swapping},
  author={Li, Lingzhi and Bao, Jianmin and Yang, Hao and Chen, Dong and Wen, Fang},
  journal={arXiv preprint arXiv:1912.13457},
  year={2019}
}

@inproceedings{frequency-aware,
  title={Frequency-aware deepfake detection: Improving generalizability through frequency space domain learning},
  author={Tan, Chuangchuang and Zhao, Yao and Wei, Shikui and Gu, Guanghua and Liu, Ping and Wei, Yunchao},
  booktitle={Proceedings of the AAAI Conference on Artificial Intelligence},
  volume={38},
  number={5},
  pages={5052--5060},
  year={2024}
}

@article{GAN,
  title={Generative adversarial nets},
  author={Goodfellow, Ian J and Pouget-Abadie, Jean and Mirza, Mehdi and Xu, Bing and Warde-Farley, David and Ozair, Sherjil and Courville, Aaron and Bengio, Yoshua},
  journal={Advances in neural information processing systems},
  volume={27},
  year={2014}
}

@article{AVSpeech,
  title={Looking to listen at the cocktail party: A speaker-independent audio-visual model for speech separation},
  author={Ephrat, Ariel and Mosseri, Inbar and Lang, Oran and Dekel, Tali and Wilson, Kevin and Hassidim, Avinatan and Freeman, William T and Rubinstein, Michael},
  journal={arXiv preprint arXiv:1804.03619},
  year={2018}
}

@article{liu2024turns,
  title={Turns Out I'm Not Real: Towards Robust Detection of AI-Generated Videos},
  author={Liu, Qingyuan and Shi, Pengyuan and Tsai, Yun-Yun and Mao, Chengzhi and Yang, Junfeng},
  journal={arXiv preprint arXiv:2406.09601},
  year={2024}
}

@inproceedings{qian2020thinking,
  title={Thinking in frequency: Face forgery detection by mining frequency-aware clues},
  author={Qian, Yuyang and Yin, Guojun and Sheng, Lu and Chen, Zixuan and Shao, Jing},
  booktitle={European conference on computer vision},
  pages={86--103},
  year={2020},
  organization={Springer}
}

@inproceedings{altfreezing,
  title={Altfreezing for more general video face forgery detection},
  author={Wang, Zhendong and Bao, Jianmin and Zhou, Wengang and Wang, Weilun and Li, Houqiang},
  booktitle={Proceedings of the IEEE/CVF conference on computer vision and pattern recognition},
  pages={4129--4138},
  year={2023}
}

@article{peng2024deepfakes,
  title={Where deepfakes gaze at? Spatial--temporal gaze inconsistency analysis for video face forgery detection},
  author={Peng, Chunlei and Miao, Zimin and Liu, Decheng and Wang, Nannan and Hu, Ruimin and Gao, Xinbo},
  journal={IEEE Transactions on Information Forensics and Security},
  volume={19},
  pages={4507--4517},
  year={2024},
  publisher={IEEE}
}

@article{li2025survey,
  title={A survey on speech deepfake detection},
  author={Li, Menglu and Ahmadiadli, Yasaman and Zhang, Xiao-Ping},
  journal={ACM Computing Surveys},
  volume={57},
  number={7},
  pages={1--38},
  year={2025},
  publisher={ACM New York, NY}
}

@article{proGAN,
  title={Progressive growing of gans for improved quality, stability, and variation},
  author={Karras, Tero and Aila, Timo and Laine, Samuli and Lehtinen, Jaakko},
  journal={arXiv preprint arXiv:1710.10196},
  year={2017}
}

@inproceedings{styleGAN,
  title={A style-based generator architecture for generative adversarial networks},
  author={Karras, Tero and Laine, Samuli and Aila, Timo},
  booktitle={Proceedings of the IEEE/CVF conference on computer vision and pattern recognition},
  pages={4401--4410},
  year={2019}
}

@inproceedings{bert,
  title={Bert: Pre-training of deep bidirectional transformers for language understanding},
  author={Devlin, Jacob and Chang, Ming-Wei and Lee, Kenton and Toutanova, Kristina},
  booktitle={Proceedings of the 2019 conference of the North American chapter of the association for computational linguistics: human language technologies, volume 1 (long and short papers)},
  pages={4171--4186},
  year={2019}
}

@article{avhubert,
  title={Learning audio-visual speech representation by masked multimodal cluster prediction},
  author={Shi, Bowen and Hsu, Wei-Ning and Lakhotia, Kushal and Mohamed, Abdelrahman},
  journal={arXiv preprint arXiv:2201.02184},
  year={2022}
}

@article{v-jepa,
  title={V-jepa 2: Self-supervised video models enable understanding, prediction and planning},
  author={Assran, Mido and Bardes, Adrien and Fan, David and Garrido, Quentin and Howes, Russell and Muckley, Matthew and Rizvi, Ammar and Roberts, Claire and Sinha, Koustuv and Zholus, Artem and others},
  journal={arXiv preprint arXiv:2506.09985},
  year={2025}
}

@article{infonce,
  title={Representation learning with contrastive predictive coding},
  author={Oord, Aaron van den and Li, Yazhe and Vinyals, Oriol},
  journal={arXiv preprint arXiv:1807.03748},
  year={2018}
}

@article{supervised-contrastive-learning,
  title={Supervised contrastive learning},
  author={Khosla, Prannay and Teterwak, Piotr and Wang, Chen and Sarna, Aaron and Tian, Yonglong and Isola, Phillip and Maschinot, Aaron and Liu, Ce and Krishnan, Dilip},
  journal={Advances in neural information processing systems},
  volume={33},
  pages={18661--18673},
  year={2020}
}

@inproceedings{clip,
  title={Learning transferable visual models from natural language supervision},
  author={Radford, Alec and Kim, Jong Wook and Hallacy, Chris and Ramesh, Aditya and Goh, Gabriel and Agarwal, Sandhini and Sastry, Girish and Askell, Amanda and Mishkin, Pamela and Clark, Jack and others},
  booktitle={International conference on machine learning},
  pages={8748--8763},
  year={2021},
  organization={PmLR}
}

@inproceedings{simCLR,
  title={A simple framework for contrastive learning of visual representations},
  author={Chen, Ting and Kornblith, Simon and Norouzi, Mohammad and Hinton, Geoffrey},
  booktitle={International conference on machine learning},
  pages={1597--1607},
  year={2020},
  organization={PmLR}
}

@inproceedings{fsfm,
  title={Fsfm: A generalizable face security foundation model via self-supervised facial representation learning},
  author={Wang, Gaojian and Lin, Feng and Wu, Tong and Liu, Zhenguang and Ba, Zhongjie and Ren, Kui},
  booktitle={Proceedings of the Computer Vision and Pattern Recognition Conference},
  pages={24364--24376},
  year={2025}
}

@inproceedings{opensdi,
  title={OpenSDI: Spotting Diffusion-Generated Images in the Open World},
  author={Wang, Yabin and Huang, Zhiwu and Hong, Xiaopeng},
  booktitle={Proceedings of the Computer Vision and Pattern Recognition Conference},
  pages={4291--4301},
  year={2025}
}

@inproceedings{cav-mae-sync,
  title={CAV-MAE Sync: Improving Contrastive Audio-Visual Mask Autoencoders via Fine-Grained Alignment},
  author={Araujo, Edson and Rouditchenko, Andrew and Gong, Yuan and Bhati, Saurabhchand and Thomas, Samuel and Kingsbury, Brian and Karlinsky, Leonid and Feris, Rogerio and Glass, James R and Kuehne, Hilde},
  booktitle={Proceedings of the Computer Vision and Pattern Recognition Conference},
  pages={18794--18803},
  year={2025}
}

@inproceedings{avf-mae++,
  title={AVF-MAE++: Scaling Affective Video Facial Masked Autoencoders via Efficient Audio-Visual Self-Supervised Learning},
  author={Wu, Xuecheng and Sun, Heli and Wang, Yifan and Nie, Jiayu and Zhang, Jie and Wang, Yabing and Xue, Junxiao and He, Liang},
  booktitle={Proceedings of the Computer Vision and Pattern Recognition Conference},
  pages={9142--9153},
  year={2025}
}

@inproceedings{crossmae,
  title={Crossmae: Cross-modality masked autoencoders for region-aware audio-visual pre-training},
  author={Guo, Yuxin and Sun, Siyang and Ma, Shuailei and Zheng, Kecheng and Bao, Xiaoyi and Ma, Shijie and Zou, Wei and Zheng, Yun},
  booktitle={Proceedings of the IEEE/CVF Conference on Computer Vision and Pattern Recognition},
  pages={26721--26731},
  year={2024}
}

@article{mavil,
  title={Mavil: Masked audio-video learners},
  author={Huang, Po-Yao and Sharma, Vasu and Xu, Hu and Ryali, Chaitanya and Li, Yanghao and Li, Shang-Wen and Ghosh, Gargi and Malik, Jitendra and Feichtenhofer, Christoph and others},
  journal={Advances in neural information processing systems},
  volume={36},
  pages={20371--20393},
  year={2023}
}

@inproceedings{whisper,
  title={Robust speech recognition via large-scale weak supervision},
  author={Radford, Alec and Kim, Jong Wook and Xu, Tao and Brockman, Greg and McLeavey, Christine and Sutskever, Ilya},
  booktitle={International conference on machine learning},
  pages={28492--28518},
  year={2023},
  organization={PMLR}
}

@misc{easyface,
    author = {Sithu Aung},
    title = {EasyFace: Easy Face Analysis Tool with SOTA Models},
    note = {Accessed: 2025-11-03},
    howpublished = {\url{https://github.com/sithu31296/EasyFace}}

}

@article{chung2018voxceleb2,
  title={Voxceleb2: Deep speaker recognition},
  author={Chung, Joon Son and Nagrani, Arsha and Zisserman, Andrew},
  journal={arXiv preprint arXiv:1806.05622},
  year={2018}
}

@inproceedings{wolf2011face,
  title={Face recognition in unconstrained videos with matched background similarity},
  author={Wolf, Lior and Hassner, Tal and Maoz, Itay},
  booktitle={CVPR 2011},
  pages={529--534},
  year={2011},
  organization={IEEE}
}

@inproceedings{gemmeke2017audio,
  title={Audio set: An ontology and human-labeled dataset for audio events},
  author={Gemmeke, Jort F and Ellis, Daniel PW and Freedman, Dylan and Jansen, Aren and Lawrence, Wade and Moore, R Channing and Plakal, Manoj and Ritter, Marvin},
  booktitle={2017 IEEE international conference on acoustics, speech and signal processing (ICASSP)},
  pages={776--780},
  year={2017},
  organization={IEEE}
}

@inproceedings{loshchilov2018decoupled,
  title={Decoupled Weight Decay Regularization},
  author={Loshchilov, Ilya and Hutter, Frank},
  booktitle={International Conference on Learning Representations},
  year={2018}
}

@inproceedings{loshchilov2016sgdr,
  title={SGDR: Stochastic Gradient Descent with Warm Restarts},
  author={Loshchilov, Ilya and Hutter, Frank},
  booktitle={International Conference on Learning Representations},
  year={2016}
}

@article{effort,
  title={Orthogonal subspace decomposition for generalizable ai-generated image detection},
  author={Yan, Zhiyuan and Wang, Jiangming and Jin, Peng and Zhang, Ke-Yue and Liu, Chengchun and Chen, Shen and Yao, Taiping and Ding, Shouhong and Wu, Baoyuan and Yuan, Li},
  journal={arXiv preprint arXiv:2411.15633},
  year={2024}
}

@inproceedings{LSDA,
  title={Transcending forgery specificity with latent space augmentation for generalizable deepfake detection},
  author={Yan, Zhiyuan and Luo, Yuhao and Lyu, Siwei and Liu, Qingshan and Wu, Baoyuan},
  booktitle={Proceedings of the IEEE/CVF Conference on Computer Vision and Pattern Recognition},
  pages={8984--8994},
  year={2024}
}

@article{poundnet,
  title={Penny-Wise and Pound-Foolish in AI-Generated Image Detection},
  author={Wang, Yabin and Huang, Zhiwu and Su, Zhou and Prugel-Bennett, Adam and Hong, Xiaopeng},
  journal={IEEE Transactions on Pattern Analysis and Machine Intelligence},
  year={2026},
  publisher={IEEE}
}

@article{dflip,
  title={Linguistic profiling of deepfakes: An open database for next-generation deepfake detection},
  author={Wang, Yabin and Hong, Xiaopeng and Li, Yaqi and Ma, Zhiheng and Huang, Zhiwu},
  journal={Pattern Recognition},
  pages={113395},
  year={2026},
  publisher={Elsevier}
}

@inproceedings{SBI,
  title={Detecting deepfakes with self-blended images},
  author={Shiohara, Kaede and Yamasaki, Toshihiko},
  booktitle={Proceedings of the IEEE/CVF conference on computer vision and pattern recognition},
  pages={18720--18729},
  year={2022}
}

@inproceedings{gu2022exploiting,
  title={Exploiting fine-grained face forgery clues via progressive enhancement learning},
  author={Gu, Qiqi and Chen, Shen and Yao, Taiping and Chen, Yang and Ding, Shouhong and Yi, Ran},
  booktitle={Proceedings of the AAAI conference on artificial intelligence},
  volume={36},
  number={1},
  pages={735--743},
  year={2022}
}

@inproceedings{sladd,
  title={Self-supervised learning of adversarial example: Towards good generalizations for deepfake detection},
  author={Chen, Liang and Zhang, Yong and Song, Yibing and Liu, Lingqiao and Wang, Jue},
  booktitle={Proceedings of the IEEE/CVF conference on computer vision and pattern recognition},
  pages={18710--18719},
  year={2022}
}

@inproceedings{yan2025generalizing,
  title={Generalizing deepfake video detection with plug-and-play: Video-level blending and spatiotemporal adapter tuning},
  author={Yan, Zhiyuan and Zhao, Yandan and Chen, Shen and Guo, Mingyi and Fu, Xinghe and Yao, Taiping and Ding, Shouhong and Wu, Yunsheng and Yuan, Li},
  booktitle={Proceedings of the Computer Vision and Pattern Recognition Conference},
  pages={12615--12625},
  year={2025}
}
}


\end{document}